\documentclass[journal]{IEEEtran}

\usepackage{amssymb}
\usepackage{amsmath}
\usepackage{booktabs}
\usepackage{multirow}
\usepackage{gensymb}

\usepackage{comment}
\usepackage{url}
\usepackage{hyperref}
\usepackage{xcolor}
\usepackage{colortbl}
\usepackage{orcidlink}
\usepackage{graphicx}
\usepackage{stfloats}

\usepackage[switch]{lineno} 
\usepackage{etoolbox}        

\newcommand*\patchAmsMathEnvironmentForLineno[1]{%
  \expandafter\pretocmd\csname #1\endcsname{\linenomath}{}{}%
  \expandafter\pretocmd\csname end#1\endcsname{\endlinenomath}{}{}%
}
\newcommand*\patchBothAmsMathEnvironmentsForLineno[1]{%
  \patchAmsMathEnvironmentForLineno{#1}%
  \patchAmsMathEnvironmentForLineno{#1*}%
}
\AtBeginDocument{%
  \patchBothAmsMathEnvironmentsForLineno{equation}%
  \patchBothAmsMathEnvironmentsForLineno{align}%
  \patchBothAmsMathEnvironmentsForLineno{gather}%
  \patchBothAmsMathEnvironmentsForLineno{multline}%
  \patchBothAmsMathEnvironmentsForLineno{flalign}%
}

\definecolor{myrefcolor}{RGB}{0,102,204}  
\definecolor{myurlcolor}{RGB}{220,50,47}  
\definecolor{mycitecolor}{RGB}{34,139,34} 

\hypersetup{
    colorlinks=true,
    linkcolor=blue,           
    citecolor=mycitecolor,     
    urlcolor=myurlcolor,      
    filecolor=magenta,
    linktoc=all
}

\title{{LiDAR, GNSS and IMU Sensor Fine Alignment through Dynamic Time Warping to Construct 3D City Maps}}

\author{%
  Haitian~Wang\,\orcidlink{0009-0000-7544-4667}\IEEEauthorrefmark{1},
  Hezam~Albaqami\,\orcidlink{0000-0003-4404-9204}\IEEEauthorrefmark{2},
  Xinyu~Wang\,\orcidlink{0009-0008-4065-2472}\IEEEauthorrefmark{1},
  Muhammad~Ibrahim,\orcidlink{0000-0002-5376-2477}\IEEEauthorrefmark{1},
  Zainy~M.~Malakan\,\orcidlink{0000-0002-6980-0992}\IEEEauthorrefmark{3},
  Abdullah~M.~Algamdi\,\orcidlink{0000-0001-9828-0335}\IEEEauthorrefmark{2},
  Mohammed~H.~Alghamdi\,\orcidlink{0000-0002-5042-0690}\IEEEauthorrefmark{4}\IEEEauthorrefmark{5},
  and~Ajmal~Mian\orcidlink{0000-0002-5206-3842}\IEEEauthorrefmark{1}%
  \thanks{Corresponding author: Hezam Albaqami (haalbaqamii@uj.edu.sa).}%
  \thanks{\IEEEauthorrefmark{1}Department of Computer Science and Software Engineering, University of Western Australia, Perth, WA 6009, Australia.}%
  \thanks{\IEEEauthorrefmark{2}Department of Computer Science and Artificial Intelligence, College of Computer Science and Engineering, University of Jeddah, Jeddah 21493, Saudi Arabia.}%
  \thanks{\IEEEauthorrefmark{3}Data Science Department, Umm Al-Qura University, Makkah 24382, Saudi Arabia.}%
  \thanks{\IEEEauthorrefmark{4}Department of Information and Technology Systems, College of Computer Science and Engineering, University of Jeddah, Jeddah 21493, Saudi Arabia.}%
  \thanks{\IEEEauthorrefmark{5}Department of Informatics and Computer Systems, College of Computer Science, King Khalid University, Abha, Saudi Arabia.}%
  \thanks{This work was funded by the University of Jeddah, Jeddah, Saudi Arabia, under grant No. (UJ-24-SUTU-1290).}%
}

\begin{document}

\maketitle

\begin{abstract}
LiDAR-based 3D mapping suffers from cumulative drift causing global misalignment, particularly in GNSS-constrained environments. To address this, we propose a unified framework that fuses LiDAR, GNSS, and IMU data for high-resolution city-scale mapping. The method performs velocity-based temporal alignment using Dynamic Time Warping and refines GNSS and IMU signals via extended Kalman filtering. Local maps are built using Normal Distributions Transform-based registration and pose graph optimization with loop closure detection, while global consistency is enforced using GNSS-constrained anchors followed by fine registration of overlapping segments. We also introduce a large-scale multimodal dataset captured in Perth, Western Australia to facilitate future research in this direction. Our dataset comprises 144{,}000 frames acquired with a 128-channel Ouster LiDAR, synchronized RTK-GNSS trajectories, and MEMS-IMU measurements across 21 urban loops. To assess geometric consistency, we evaluated our method using alignment metrics based on road centerlines and intersections to capture both global and local accuracy. The proposed framework reduces the average global alignment error from 3.32\,m to 1.24\,m, achieving a 61.4\% improvement, and significantly decreases the intersection centroid offset from 13.22\,m to 2.01\,m, corresponding to an 84.8\% enhancement. {The constructed high-fidelity map and raw dataset are publicly available through \href{https://ieee-dataport.org/documents/perth-cbd-high-resolution-lidar-map-gnss-and-imu-calibration}{IEEE Dataport} and its visualization can be viewed in the provided \href{https://www.youtube.com/watch?v=-ZUgs1KyMks}{Demo}. This dataset and method together establish a new benchmark for evaluating 3D city mapping in GNSS-constrained environments, with source code available at \href{https://github.com/HaitianWang/LiDAR-GNSS-and-IMU-Sensor-Fine-Alignment-through-Dynamic-Time-Warping-to-Construct-3D-City-Maps}{GitHub Repository}.}
\end{abstract}


\vspace{-2mm}
\section{Introduction}
\label{Introduction}

Urbanization is rapidly transforming cities into dense and complex environments, increasing the demand for scalable infrastructure planning and maintenance~\cite{brown2023regulating, lim2024australia}. In this context, updated high-resolution spatial data is essential~\cite{yin2021rall, koide2021globally, ibrahim2025forest}. However, passive vision-based methods such as photogrammetry~\cite{jiang2021unmanned}, stereo vision~\cite{esparza2022stdyn}, and Structure-from-Motion (SfM)~\cite{deliry2021accuracy} often struggle in poor or changing lighting conditions, making them unsuitable for detailed city-scale 3D mapping.

\begin{figure*}[!t]
    \centering
    \includegraphics[width=\textwidth]{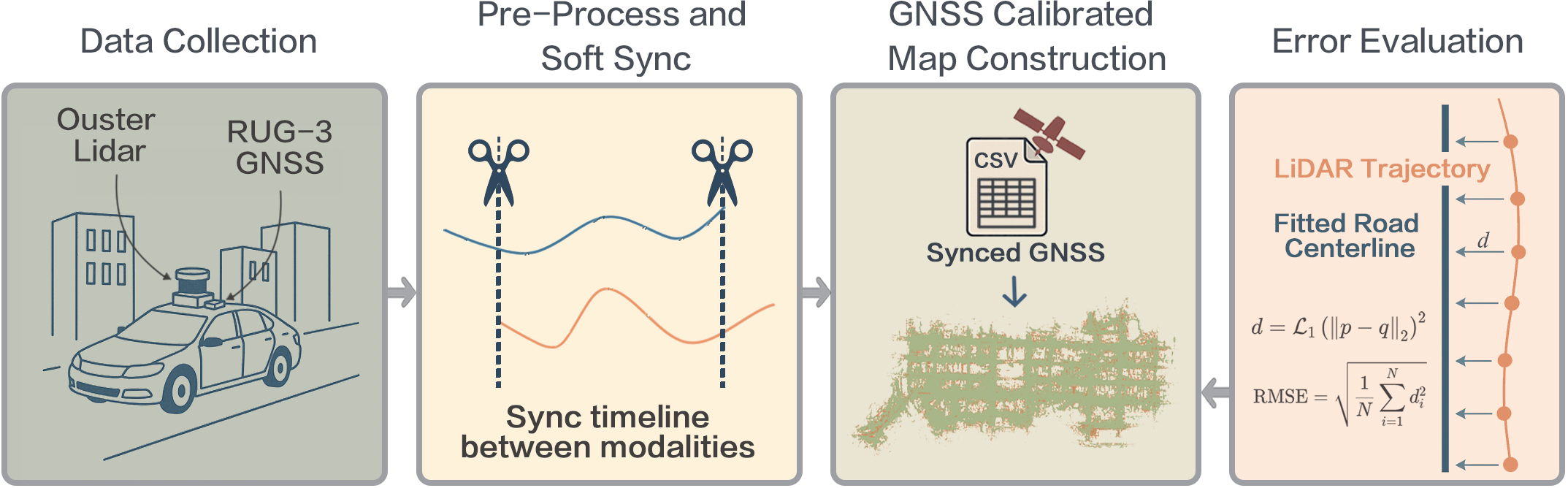}
    \caption{Overview of the proposed framework for 3D city mapping which comprises: (1) multimodal data capture via Ouster OS1-128 LiDAR and RUG-3 GNSS/IMU; (2) preprocessing with velocity-based alignment and Dynamic Time Warping synchronization; (3) GNSS-calibrated 3D mapping using NDT, pose graph optimization, and global ICP; and (4) accuracy evaluation via skeleton-based centreline metrics against 2D cartographic references.}
    \label{fig:overview-diagram}
    \vspace{-6mm}
\end{figure*}

Light Detection and Ranging (LiDAR) sensor provides dense 3D data with rich geometric features that are robust to lighting and weather conditions, making it ideal for mapping complex urban environments~\cite{yu2025slim}. LiDAR can accurately capture features such as curbs, intersections, poles, road markings, traffic signs, building facades, sidewalks, overpasses, and street furniture. These capabilities make it a key enabler of precise city-scale 3D mapping. Recent advances, including the integration of solid-state LiDAR and IMUs~\cite{ma2024solid}, fusion with high-precision GNSS~\cite{li2018high, ruwisch2025feature, 2025lidar}, and digital LiDAR architectures such as Ouster multibeam sensors~\cite{ouster2021digital} have contributed to the advancement of 3D city mapping in dense urban environments~\cite{liu2023ubiquitous, du2023gnss, wen20213d}.

Despite these advancements, LiDAR-based mapping still faces challenges such as cumulative drift during frame-by-frame alignment~\cite{he2022lidar, di2021mobile}, especially over long trajectories. Although methods like In2Laama~\cite{le2020in2laama} address motion distortion through LiDAR-IMU optimization, they lack GNSS integration and do not scale well for large urban deployments. GNSS can help mitigate drift, but in dense cityscape, signals are often degraded or lost due to occlusions by tall buildings, bridges, or tunnels. Moreover, many LiDAR platforms do not include native GNSS receivers, and external units require precise calibration and synchronization. These limitations lead to misalignments that compromise the global consistency and reliability of the final map.

{Two representative approaches to multimodal sensor integration are batch spatiotemporal calibration using continuous-time factor graphs and latent-space sequence alignment using Canonical Time Warping. The first approach jointly estimates a continuous-time trajectory and a single inter-sensor delay parameter from camera, LiDAR, and IMU factors \cite{rehder2016tro}. These formulations assume a constant offset during a session and require sufficient six-degree-of-freedom excitation to render the time parameter observable, which is often violated in long urban drives with limited roll and pitch change. The normal equations grow with the number of knots and measurements, which limits scalability on city-scale logs. Moreover, continuous-time LiDAR odometry corrects intra-scan motion distortion but does not align independent IMU and GNSS clocks when the delay is nonstationary over time \cite{vizzo2022cticp, 2025citymultistream, 2024high}. The second approach, including Canonical Time Warping and its deep learning variants, learns projections that map two sequences into a shared latent space and then performs alignment in that space \cite{zhou2009ctw,trigeorgis2016dctw, wang2024city}. These methods are effective for high-dimensional paired observations such as speech features or articulated motion, yet LiDAR–IMU synchronization provides low-dimensional kinematic cues like speed and heading, cross-view correlations degrade under long stops and sensor dropouts, and learning the projections increases memory and runtime on long urban loops when task-specific training data are unavailable.}

These sensor-level limitations are further compounded by gaps in existing urban mapping datasets. Widely used benchmarks such as KITTI~\cite{liao2022kitti, karri2023key, behley2019semantickitti}, Perth-WA~\cite{s2p2-2e66-23} and nuScenes~\cite{caesar2020nuscenes} lack reliable GNSS ground truth and globally consistent 3D map alignment. This makes them unsuitable for evaluating city-scale mapping in environments where GNSS signals are degraded. Moreover, datasets such as Toronto-3D~\cite{tan2020toronto} and the Melbourne Urban Dataset~\cite{gupta2019tree}, offer improved spatial coverage but still suffer from sparse road-level detail, frequent occlusions, and reduced accuracy in dense areas. To address some of these challenges, various multi-sensor fusion frameworks~\cite{turn0search0, turn0search1, turn0search6} and semantic-aided stabilization methods~\cite{zhao2021fusion, he2022urban, ibrahim2025forest} have been developed. However, most are limited to short-ranges and fail to correct cumulative drift or achieve accurate synchronization across modalities in large-scale, GNSS-constrained settings. Furthermore, robust temporal synchronization across heterogeneous sensors remains largely unexplored in existing datasets and methods.

To overcome the challenges of cumulative drift and cross-sensor synchronization in city-scale mapping, we propose LIGMA (LiDAR-IMU-GNSS Mapping Architecture), a unified multimodal fusion framework that integrates data from the three sensors for globally consistent 3D reconstruction. The core novelty lies in a Dynamic Time Warping (DTW)-based velocity matching strategy that enables robust temporal alignment across sensors with varying sampling rates and unsynchronized clocks. To our knowledge, this is the first use of DTW-based synchronization in urban LiDAR-GNSS mapping. Our method further employs a hierarchical registration approach that includes local alignment via Normal Distributions Transform (NDT), pose graph optimization with loop closure detection, and global refinement using Iterative Closest Point (ICP). GNSS and IMU measurements serve as spatial anchors, mitigating cumulated drift to improve global consistency.

Alongside the proposed framework, we present a high-resolution multisensor dataset collected in the Perth Central Business District (CBD), Western Australia. The dataset comprises 144,000 LiDAR frames from a 128-channel Ouster sensor, accompanied by synchronized RTK-GNSS and IMU measurements across 21 road loops, covering approximately the $4.2,\text{km}^2$ region. LiDAR data are preprocessed through Statistical Outlier Removal (SOR) for denoising and voxel grid filtering for downsampling. GNSS and IMU streams are processed with an Extended Kalman Filter and temporally aligned via DTW to ensure consistency across modalities. This dataset provides dense 3D geometry with precise georeferencing and supports benchmarking of city mapping in GNSS-denied or obstructed environments.
{
DTW performs well when the motion signal contains stop–go events and speed transients, but the cost surface becomes ambiguous under near-constant speed or prolonged stops. We therefore restrict warping with a Sakoe–Chiba band and seed stationary anchors; when the windowed speed variance is low we fall back to a constant-shift estimate.
}

To evaluate geometric consistency, we employ a dual-metric approach using both centreline-based and intersection-based alignment metrics, providing a comprehensive assessment of global and local spatial accuracy. Specifically, our method reduces the global centreline root mean square error (RMSE) from 3.32,m to 1.24,m, representing a 61.4\% improvement, and decreases the mean intersection centroid offset from 13.22,m to 2.01,m, achieving an 84.8\% enhancement in local geometric fidelity. The complete dataset, consisting of 144,000 LiDAR frames with synchronized RTK-GNSS and MEMS-IMU measurements across 21 urban loops in Perth CBD, is publicly available at \href{https://ieee-dataport.org/documents/perth-cbd-high-resolution-lidar-map-gnss-and-imu-calibration}{IEEE Dataport}, with an interactive demonstration accessible at \href{https://www.youtube.com/watch?v=-ZUgs1KyMks}{Demo}. The modular design of the proposed framework facilitates adaptation to varied sensor configurations and urban topologies, supporting its application in diverse large-scale mapping scenarios and serving as a benchmark for future research in GNSS-challenged urban environments.

{Our contributions are summarized below: 
\begin{enumerate}
     \item We propose a method for constructing 3D city maps using non-synchrous LiDAR, GNSS and IMU data and its visualization can be viewed in the provided \href{https://www.youtube.com/watch?v=-ZUgs1KyMks}{Demo}.
     \item We propose a unified sensor fusion pipeline, coined LIGMA, that integrates LiDAR, GNSS, and inertial data using a novel velocity-based alignment and multi-stage registration strategy. Source code has been available at \href{https://github.com/HaitianWang/LiDAR-GNSS-and-IMU-Sensor-Fine-Alignment-through-Dynamic-Time-Warping-to-Construct-3D-City-Maps}{GitHub Repository}.
     \item We present a high-resolution dataset captured with 128 channel Ouster LiDAR covering $4.2\,\text{km}^2$ of central Perth in Western Australia. The dataset comprises 144,000 LiDAR frames along with synchronized and georeferenced inertial and positioning data. The constructed high-fidelity map and raw dataset are publicly available through \href{https://ieee-dataport.org/documents/perth-cbd-high-resolution-lidar-map-gnss-and-imu-calibration}{IEEE Dataport}
\end{enumerate}}

\vspace{-4mm}
\section{Data Collection}
\label{Data Collection}
\vspace{-2mm}
This section outlines the data acquisition framework, covering survey objectives, sensor setup, deployment strategy, data formats, and quality assurance. The goal is to ensure high geospatial accuracy, temporal consistency, and broad urban coverage. The resulting dataset supports downstream tasks including multi-sensor 3D mapping, localization, and change detection.

\begin{figure*}[!t]
\centering
\includegraphics[width=\textwidth]{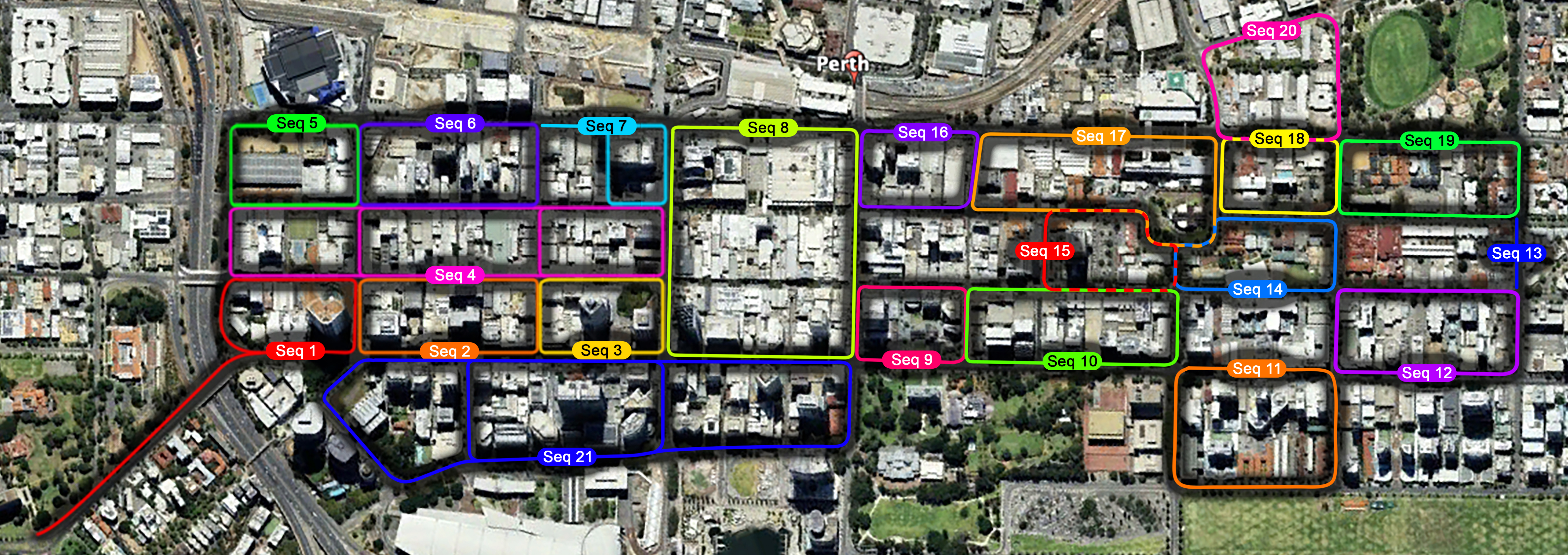}
\caption{Survey layout of the Perth CBD dataset with 21 manually designed loops (Seq 1–21) spanning 18.6km of urban roads, providing diverse viewpoints, repeated coverage, and intersection overlap for robust SLAM and high-quality 3D mapping.}
\label{fig:collection-plan}
\vspace{-6mm}
\end{figure*}

\vspace{-3mm}
\subsection{Survey Objectives and Study Area Definition}
This survey aims to construct a high-precision, multi-sensor 3D point cloud dataset of Perth CBD to support urban mapping, planning, and large-scale monitoring. It also provides a foundation for developing and evaluating algorithms for LiDAR-based SLAM, GNSS–LiDAR fusion, and change detection. Perth CBD was selected for its structural complexity, frequent GNSS occlusions, dynamic traffic, and variable weather—conditions that pose significant challenges for localization, perception, and long-term mapping.

The dataset captures a wide range of urban features, including high-rise buildings, multi-lane roads, laneways, sidewalks, intersections, pedestrian zones, green corridors, and public infrastructure such as signage, transit stops, and light poles. To ensure spatial coverage and temporal redundancy, data collection was structured into 21 manually designed loops totaling 18.6\,km, with an average loop length of 931\,m. Frequent intersections among loops facilitate loop closure detection and improve spatial consistency. The routes span major and minor roads, one-way streets, and complex junctions, ensuring diverse scene capture from multiple perspectives. As shown in Fig.~\ref{fig:collection-plan}, the survey layout enables repeated traversals and varied viewpoints for robust SLAM and map refinement. Data were collected over several days using a vehicle traveling at 10–30\,km/h during low-traffic periods to ensure high point density and minimize motion distortion and occlusion.

\begin{figure*}[!b]
\centering
\includegraphics[width=\textwidth]{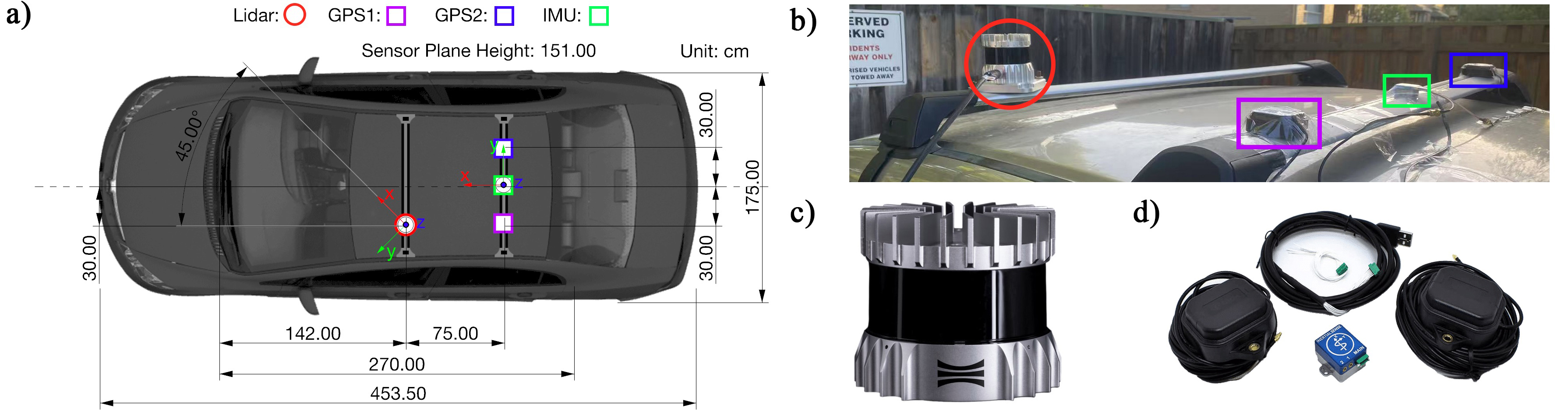}
\vspace{-2mm}
\caption{Sensor setup for vehicle-based data collection in central Perth, featuring a roof-mounted 128-channel LiDAR (red), dual GNSS antennas (purple and blue), and a high-grade inertial unit (green), all precisely aligned. (a) Schematic with dimensions, (b) real-world deployment, and (c--d) core sensor modules.}
\label{fig:sensor-setup-configuration}
\end{figure*}

\vspace{-2mm}
\subsection{Sensor Specifications}
The sensing system comprises a 128-channel Ouster OS1 LiDAR and a RUG-3-IMX-5-Dual GNSS–IMU module~\cite{inertialsense_rug3}. The OS1 is a dual-return high-resolution scanner with a 45\degree{} vertical field of view, 0.35\degree{} angular resolution at 10~Hz, and a peak output of 2.6 million points per second. Each point includes 3D coordinates, intensity, and a time stamp, which supports detailed reconstruction of urban environments~\cite{ouster_datasheet,ouster_usermanual}. The GNSS–IMU module integrates dual multi-frequency RTK receivers and a high-performance MEMS inertial unit, delivering sub-meter accuracy and full 6-DoF orientation. It supports GPS, GLONASS, Galileo, and BeiDou constellations. According to the manufacturer specifications, the GNSS operates at 5~Hz and the IMU at 73~Hz~\cite{inertialsense_usermanual}. Unlike tightly coupled systems, the LiDAR and GNSS–IMU modules operate asynchronously, with synchronization and logging handled by a central acquisition system as detailed in subsection~\ref{Sensor_Setup and Configuration}.

\vspace{-2mm}
\subsection{Sensor Setup and Hardware Configuration}
\label{Sensor_Setup and Configuration}
The sensor suite was mounted on a standard passenger vehicle with careful attention to spatial alignment and mechanical stability. As shown in Fig.~\ref{fig:sensor-setup-configuration}, the setup includes an Ouster LiDAR, dual GNSS antennas, and a GNSS–IMU module, all co-located and geometrically referenced on the roof. The LiDAR was positioned at the front centerline, 1.51\,m above ground, offering a clear 360\degree{} horizontal field of view. GNSS antennas were rigidly mounted at the front-left and rear-right corners, forming a 1.42\,m baseline for yaw estimation. The IMU was placed centrally between the antennas along the vehicle's longitudinal axis to minimize lever-arm effects and rotational distortion. Temporal synchronization was achieved via a PPS signal from the GNSS receiver, ensuring alignment across sensor streams. Data were logged using manufacturer SDKs: Ouster Sensor SDK for LiDAR and Inertial Sense uINS SDK for GNSS–IMU data. Both sensor modules operate independently and connect to a ruggedized Ubuntu 20.04 laptop: the LiDAR via Gigabit Ethernet and the GNSS–IMU through USB. All data streams are timestamped in real-time using the host clock. The devices were powered by an automotive source 12\,V and connected through industrial-grade USB and serial interfaces. Sensor extrinsics were calibrated using manufacturer specifications and verified through constrained-motion routines. All components were housed in weather-resistant enclosures with shielded cable routing to mitigate EMI and vibration. 

{Despite PPS gating, we observed non-zero residuals between stream timestamps. Over the 21 sequences, the per-frame absolute time error between LiDAR packet timestamps and the nearest IMU epoch (after PPS+NMEA configuration) had a median of 18.7\,ms and a 95th percentile of 42.3\,ms; occasional bursts of $\approx$100\,ms coincided with host OS scheduling spikes. 
Because of these issues, we proposed: (i) LiDAR UDP buffering/packetization prior to delivery to the host, (ii) host-side timestamping latency, (iii) serial NMEA message emission phase relative to PPS, and (iv) kernel scheduling jitter. These are consistent with the Ouster timing model (common ns-resolution timer synchronized via \texttt{SYNC\_PULSE\_IN} and optional NMEA) and uINS timing characteristics (sub-$\mu$s PPS jitter, millisecond-level IMU signal latency) \cite{ouster_time_sync,inertialsense_imx5}. Quantifying these residuals motivated an explicit post-PPS temporal alignment stage.
}

\begin{figure*}[tb!]
\centering
\includegraphics[width=\textwidth]{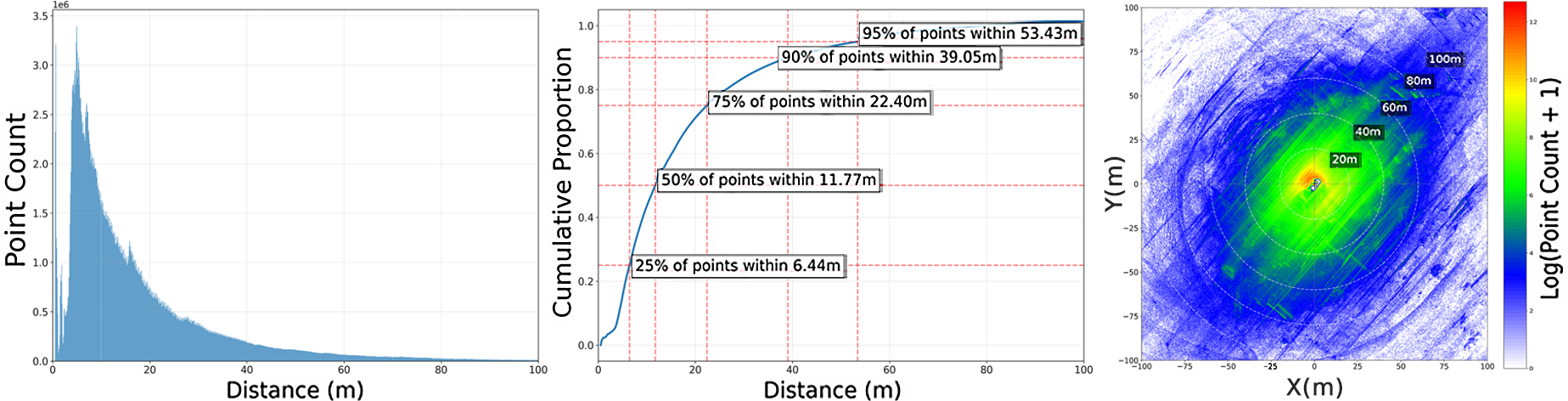}
\vspace{-2mm}
\caption{Point cloud distribution and density - Left: Point count histogram vs. distance from LiDAR origin; Middle: Cumulative distribution showing 75\% of points within 22.4m; Right: Top-down heatmap with log-scaled point density, highlighting near-field concentration.}
\label{fig:raw-data-description}
\vspace{-6mm}
\end{figure*}

\begin{figure*}[t]
    \centering
    \includegraphics[width=\textwidth, height=0.85\textwidth]{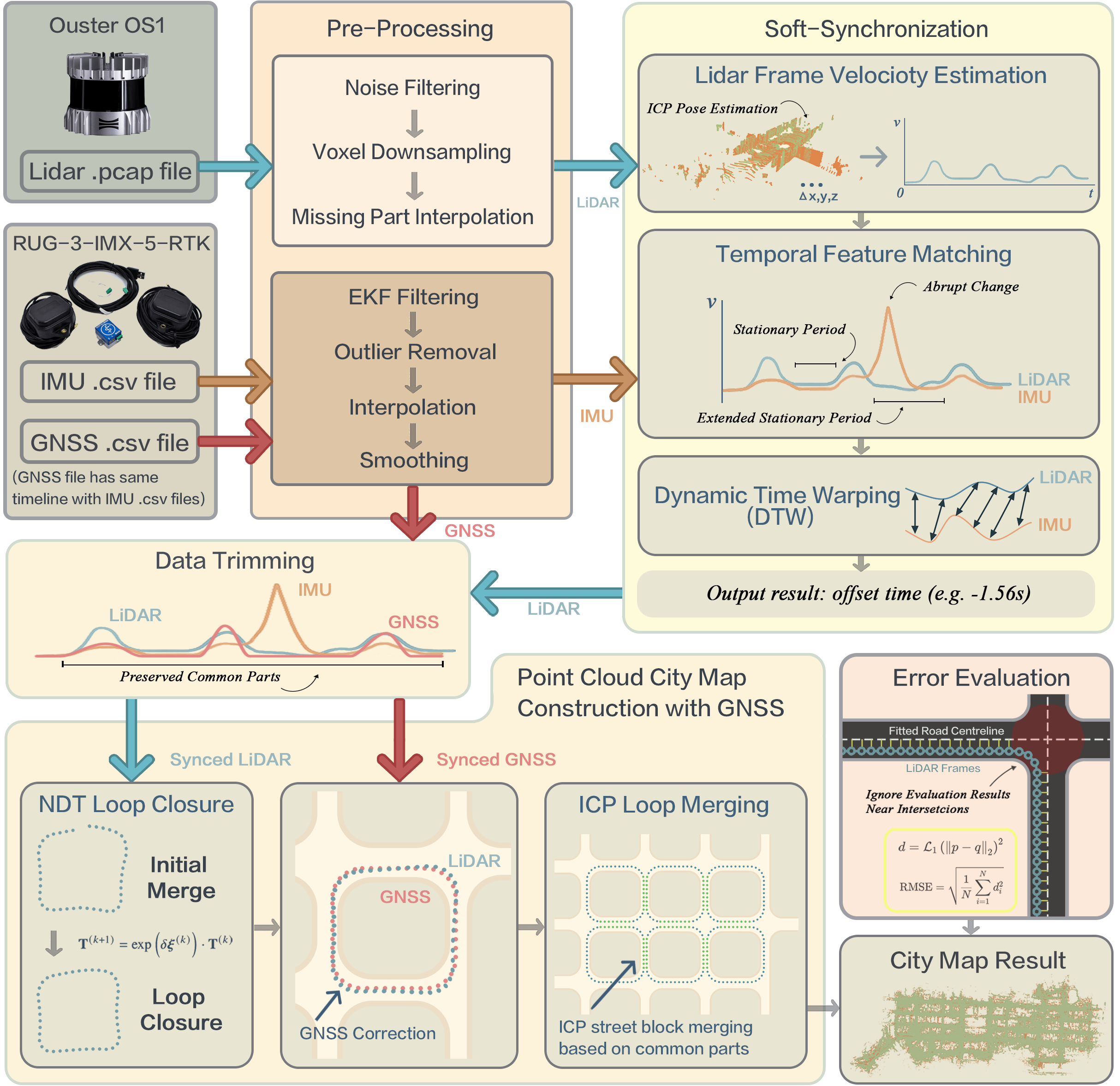}
    \vspace{-2mm}
    \caption{Overview of the proposed LIGMA pipeline for urban LiDAR–GNSS 3D map construction. The method comprises four stages: (1) LiDAR and GNSS/IMU preprocessing via denoising, downsampling, and temporal interpolation; (2) soft synchronization using velocity estimation, feature matching, and Dynamic Time Warping; (3) multimodal map construction through NDT registration, GNSS-constrained correction, and ICP-based merging; and (4) quantitative evaluation using centerline and intersection-level alignment metrics.}
    \label{fig:method-pipeline}
    \vspace{-4mm}
\end{figure*}

\vspace{-2mm}
\subsection{Data Modalities, Formats, and Organization}
The dataset is organized in a loop-based directory structure (e.g., \texttt{/seq01/}, \texttt{/seq02/}), with each sequence containing subfolders for LiDAR, GNSS, IMU, calibration, and logs. LiDAR frames were recorded at 10~Hz in dual-return mode using the Ouster OS1-128, producing approximately 380~GB of raw data. Each frame contains 260k–280k points with 3D coordinates, intensity, and ring index, initially saved in \texttt{.pcap} format and later converted to \texttt{.bin} and \texttt{.ply} for downstream use. GNSS and IMU data were logged using the Inertial Sense uINS SDK at 5~Hz and 73~Hz, respectively, and stored in \texttt{.csv} format with synchronized Unix timestamps and metadata (e.g., RTK status, satellite visibility). All sensor streams are time-aligned using PPS synchronization and spatially registered via a rigid-body extrinsic matrix defined in \texttt{transforms.yaml}. To support benchmarking, the dataset also includes pose graphs, KITTI-format trajectories, and loop closure ground truth. As shown in Fig.~\ref{fig:raw-data-description}, point density analysis indicates that 75\% of LiDAR returns lie within 22.4\,m, emphasizing the importance of high-fidelity short-range sensing for urban mapping.

\vspace{-2mm}
\subsection{Quality Assurance and Handling of Imperfections}
A multi-stage quality assurance pipeline was implemented to ensure data reliability for SLAM, sensor fusion, and urban perception tasks. During collection, real-time diagnostics monitored LiDAR return rates, satellite visibility (SV count), IMU status, and RTK fix quality via onboard dashboards. Sequences with degraded signals (e.g., SV count $< 6$ or RTK loss) were flagged for review. Data were captured over multiple low-traffic days to minimize dynamic occlusions. Post-processing included statistical and radius-based filtering to clean LiDAR noise and sparsity, while frames with excessive dropout ($>25\%$) were annotated for exclusion. GNSS trajectories were smoothed using an extended Kalman filter to correct jump artifacts. Temporal alignment was verified by comparing LiDAR odometry with GNSS–IMU dead reckoning, with sequences exceeding 0.5\,m residual or 200\,ms lag corrected or removed. Calibration was validated through rigid-body residuals and loop closure consistency; loops with drift over 0.3\,m were reprocessed or discarded. All quality flags and exclusion masks are stored in YAML metadata for selective processing and robustness evaluation.


\vspace{-2mm}
\section{Methods}
\label{Methods}

This section presents the LIGMA framework for high-precision 3D urban mapping. It leverages tightly coupled integration of LiDAR, GNSS, and IMU data. As shown in Fig.~\ref{fig:method-pipeline}, the proposed method comprises four main stages: (1) multimodal preprocessing, including noise filtering, voxel downsampling, interpolation, and state estimation via an Extended Kalman Filter; (2) temporal alignment of LiDAR and IMU streams using velocity-based matching and Dynamic Time Warping; (3) hierarchical mapping through NDT registration, GNSS-constrained loop closure, and ICP refinement; and (4) quantitative evaluation using skeleton-based centerline extraction and geometric metrics such as RMSE and Hausdorff distance. The following subsections describe each stage in detail.
{
We adopt a loosely coupled design: GNSS contributes as soft factors in the pose graph, while temporal alignment is handled in preprocessing. This keeps the synchronizer independent of GNSS availability in urban canyons and simplifies deployment at scale. Tightly coupled LiDAR–IMU–GNSS back-ends can yield further improvements when raw measurements are accessible, but they require different interfaces and formulations \cite{shan2020liosam,cioffi2020tightly}.
}

\vspace{-3mm}
\subsection{Multimodal Sensor Data Preprocessing}
Preprocessing is critical for ensuring data quality and enabling effective multimodal integration. The pipeline (Fig.~\ref{fig:method-pipeline}) handles data from the Ouster OS1-128 LiDAR and the Inertial Sense RUG-3 IMX-5 GNSS/IMU module, applying domain-specific routines for each modality.

\subsubsection{LiDAR Data}
Raw LiDAR scans are decoded using the Ouster SDK \cite{ouster_sdk_docs}, yielding point cloud frames with 3D coordinates, intensity, reflectivity, and ring ID attributes \cite{ouster_datasheet}. To enhance data quality, spurious returns from reflective or transparent surfaces are removed using statistical outlier filtering, which discards points deviating more than one standard deviation ($z=1.0$) from the mean distance within a $k$-nearest neighborhood ($k=50$)~\cite{zhou2024dcor}. For efficiency and spatial consistency, voxel grid downsampling with a 0.05~m resolution is applied, replacing all points in each voxel with their centroid~\cite{borges2022fractional, liu2024pointpillars}. Frames missing due to occlusion or transmission loss are reconstructed by linear interpolation between adjacent frames~\cite{chen2024filling}. The processed point clouds are stored in \texttt{.ply} format with aligned timestamps for multimodal integration.

\subsubsection{GNSS Data}
GNSS data sampled at 5~Hz are preprocessed to ensure trajectory reliability in complex urban environments.
{As shown in Fig.~\ref{fig:method-pipeline}, we employ an Extended Kalman Filter (EKF) as a preprocessing step to (i) smooth the GNSS trajectory under multipath and intermittent RTK degradation and (ii) provide a bias-aware inertial dead-reckoning estimate during short GNSS gaps. The EKF is implemented in a local ENU navigation frame and is propagated at the IMU rate (73~Hz) with GNSS updates applied at 5~Hz.}
EKF smoothing is applied to estimate the state vector, which includes position, velocity, orientation, and sensor biases~\cite{singh2023inverse}.

{Specifically, the EKF state at epoch $k$ is}
{
\begin{equation}
\mathbf{x}_k=\Big[\mathbf{p}_k^\top,\ \mathbf{v}_k^\top,\ \mathbf{q}_k^\top,\ \mathbf{b}_{g,k}^\top,\ \mathbf{b}_{a,k}^\top\Big]^\top,
\end{equation}
}
{where $\mathbf{p}_k\in\mathbb{R}^3$ and $\mathbf{v}_k\in\mathbb{R}^3$ are position and velocity in ENU, $\mathbf{q}_k$ is the unit quaternion representing attitude (body-to-ENU), and $\mathbf{b}_{g,k}$ and $\mathbf{b}_{a,k}$ are gyroscope and accelerometer biases. Let $\boldsymbol{\omega}_k$ and $\mathbf{a}_k$ denote the raw IMU angular rate and acceleration at time step $\Delta t$. The discrete-time propagation follows the standard strapdown mechanization:}
{
\begin{align}
\mathbf{p}_{k+1} &= \mathbf{p}_k + \mathbf{v}_k \Delta t
+ \tfrac{1}{2}\Big(\mathbf{R}(\mathbf{q}_k)\big(\mathbf{a}_k-\mathbf{b}_{a,k}\big)-\mathbf{g}\Big)\Delta t^2, \\
\mathbf{v}_{k+1} &= \mathbf{v}_k
+ \Big(\mathbf{R}(\mathbf{q}_k)\big(\mathbf{a}_k-\mathbf{b}_{a,k}\big)-\mathbf{g}\Big)\Delta t, \\
\mathbf{q}_{k+1} &= \mathbf{q}_k \otimes \exp\!\Big(\tfrac{1}{2}\big(\boldsymbol{\omega}_k-\mathbf{b}_{g,k}\big)\Delta t\Big), \\
\mathbf{b}_{g,k+1} &= \mathbf{b}_{g,k} + \mathbf{n}_{bg,k}, \qquad
\mathbf{b}_{a,k+1} = \mathbf{b}_{a,k} + \mathbf{n}_{ba,k},
\end{align}
}
{where $\mathbf{R}(\mathbf{q}_k)$ converts a quaternion to a rotation matrix, $\mathbf{g}=[0,0,9.81]^\top$\,m/s$^2$ in ENU, $\otimes$ denotes quaternion multiplication, and $\exp(\cdot)$ maps an incremental rotation vector to a unit quaternion. The bias terms follow a random-walk model with driving noise $\mathbf{n}_{bg,k}$ and $\mathbf{n}_{ba,k}$. The covariance is propagated as}
{
\begin{equation}
\mathbf{P}_{k+1|k}=\mathbf{F}_k\,\mathbf{P}_{k|k}\,\mathbf{F}_k^\top+\mathbf{Q}_k,
\end{equation}
}
{where $\mathbf{F}_k$ is the linearized state transition and $\mathbf{Q}_k$ is constructed from the IMU noise characteristics provided by the sensor/SDK configuration (gyro/accel noise and bias random-walk).}

{For the measurement update, we use the GNSS position and Doppler-derived velocity reported by the uINS solution. The observation is}
{
\begin{equation}
\mathbf{z}_k=\Big[(\mathbf{p}^{\text{GNSS}}_k)^\top,\ (\mathbf{v}^{\text{GNSS}}_k)^\top\Big]^\top,\qquad
h(\mathbf{x}_k)=\Big[\mathbf{p}_k^\top,\ \mathbf{v}_k^\top\Big]^\top,
\end{equation}
}
{with a standard EKF update $\hat{\mathbf{x}}_{k|k}=\hat{\mathbf{x}}_{k|k-1}+\mathbf{K}_k(\mathbf{z}_k-h(\hat{\mathbf{x}}_{k|k-1}))$ and $\mathbf{K}_k=\mathbf{P}_{k|k-1}\mathbf{H}^\top(\mathbf{H}\mathbf{P}_{k|k-1}\mathbf{H}^\top+\mathbf{R}_k)^{-1}$, where $\mathbf{H}$ is the Jacobian of $h(\cdot)$ and $\mathbf{R}_k$ is taken from the GNSS-reported covariance when available.}

To improve accuracy, outlier removal is performed using Mahalanobis distance thresholding, which excludes GNSS measurements with a distance greater than 3.0, often caused by multipath effects or weak satellite geometry~\cite{wen2021factor, dreissig2023survey}. For brief GNSS outages (less than one second), interpolation and smoothing are applied using cubic Hermite splines to fill gaps, followed by a low-pass Butterworth filter to suppress high-frequency noise~\cite{erol2021comparative}. The resulting data is saved in synchronized \texttt{.csv} files containing cleaned position, velocity, and RTK status.

\vspace{-1mm}
\subsubsection{IMU Data}
The IMU provides measurements of angular velocity and linear acceleration. The raw IMU data are corrected and integrated using a quaternion-based inertial navigation algorithm to estimate orientation and translational velocity. Quaternions offer numerically stable roll–pitch–yaw estimation and avoid singularities such as gimbal lock. Temporal alignment is achieved using a Pulse-Per-Second (PPS) signal from the GNSS receiver. To match the 10~Hz sampling rate of the LiDAR and the 5~hz of GNSS streams, the 73~Hz IMU data are downsampled by selecting, for each LiDAR frame, the temporally closest IMU reading. This approach minimizes skew while preserving high-frequency motion characteristics at the resolution required for sensor fusion. The calibrated motion and orientation outputs are stored in a unified \texttt{.csv} format for downstream multimodal processing.  {The GNSS and IMU measurements are produced by the same RUG-3 IMX-5 unit and are hardware timestamped on a single PPS clock. Their mutual alignment is therefore direct and does not require DTW. DTW is applied only between the OS1-128 LiDAR stream (10 Hz) and IMU-derived kinematics (73 Hz).}

\vspace{-2mm}
\subsection{Multimodal Sensor Data Synchronization}

Accurate temporal alignment is essential for multimodal fusion. LIGMA performs synchronization by matching velocity signatures derived independently from LiDAR and IMU data (Fig.~\ref{fig:method-pipeline}).

\subsubsection{LiDAR Velocity Estimation}
To estimate LiDAR velocity, consecutive point cloud frames are registered using ICP algorithm, which yields a 4$\times$4 transformation matrix representing relative motion. The translational displacement $(\Delta x, \Delta y, \Delta z)$ extracted from this matrix is used to compute instantaneous speed as $v_{\text{LiDAR}}(t) = \frac{\sqrt{\Delta x^2 + \Delta y^2 + \Delta z^2}}{\Delta t}$, where $\Delta t$ is the time interval between frames~\cite{haas2023velocity}.  {We deliberately avoid using RTK–GNSS 3D velocity for temporal alignment because in dense urban blocks the RTK status often toggles (FIX/FLT) and the 5 Hz velocity stream exhibits short gaps and smoothing latency, whereas LiDAR- and IMU-derived speed are continuously available at higher rates. Separately, keeping synchronization independent of GNSS (which is later used for global geo-anchoring) prevents data leakage and avoids optimistic bias.}
{Note that RTK-GNSS 3D velocity is not used for DTW because in dense urban canyons FIX/FLT transitions, multipath and short outages introduce rate and latency inconsistencies~\cite{fan2019precise, wang2025georegistrationterrestriallidarpoint}.  GNSS is retained only as a global geo-anchor.  Yaw is informative but enters the DTW cost as a low-weight auxiliary term, while scalar speed remains the primary alignment cue~\cite{groves2011shadow}.}

\subsubsection{IMU Velocity Estimation}
IMU-based velocity estimation involves correcting the raw acceleration measurements $\mathbf{a}_{\text{meas}}(t)$ by removing the bias $\mathbf{b}_a$ and accounting for gravity $\mathbf{g}$. The corrected accelerations are then transformed to the global frame using the rotation matrix $\mathbf{R}(t)$ derived from quaternion-based orientation, yielding $\mathbf{a}_{\text{corr}}(t) = \mathbf{R}(t)\left(\mathbf{a}_{\text{meas}}(t) - \mathbf{b}_a\right) - \mathbf{g}$ as the input for global-frame velocity integration.

This corrected acceleration is integrated over time to obtain global velocity $\mathbf{v}(t)$, and the IMU speed is computed as its Euclidean norm:
\vspace{-4mm}
\begin{equation}
\vspace{-2mm}
v_{\text{IMU}}(t) = \left\| \mathbf{v}(t-\Delta t) + \int_{t-\Delta t}^{t} \mathbf{a}_{\text{corr}}(\tau)\,d\tau \right\|_2
\vspace{2mm}
\end{equation}
where $\Delta t$ is the time step. This approach ensures robust and drift-reduced velocity estimates for multimodal synchronization and mapping.

\subsubsection{{Linear Time-Shift Baseline}}
{
We estimate a time linear offset $\hat{\tau}$ by maximizing the normalized cross-correlation between speed profiles $v_{\text{LiDAR}}(t)$ and $v_{\text{IMU}}(t)$:
\begin{equation}
\hat{\tau}=\arg\max_\tau \frac{\sum_t \bigl(v_{\text{LiDAR}}(t)-\bar v_L\bigr)\,\bigl(v_{\text{IMU}}(t+\tau)-\bar v_I\bigr)}{\sigma_L\,\sigma_I}
\end{equation}
then align by $\tilde v_{\text{IMU}}(t)=v_{\text{IMU}}(t+\hat{\tau})$. This baseline is appropriate when the timing error is well-approximated by a constant offset \cite{knapp1976}.
}

\subsubsection{{DTW-based Synchronization}}
{
Let $v_L[i]$ and $v_I[j]$ denote LiDAR- and IMU-derived speed sequences resampled at 10~Hz, and let $\psi_L[i]$ and $\psi_I[j]$ be the corresponding headings (IMU yaw projected to the LiDAR frame). We align them using Dynamic Time Warping (DTW) with a Sakoe–Chiba band $|i-j|\le W$ (we use $W=25$, i.e., 2.5\,s at 10\,Hz) and a speed+heading cost:
\begin{equation}
c(i,j)=\bigl(v_L[i]-v_I[j]\bigr)^2+\lambda\,\operatorname{wrap}_\pi\!\bigl(\psi_L[i]-\psi_I[j]\bigr)^2,
\end{equation}
where $\operatorname{wrap}_\pi(\cdot)$ maps angles to $(-\pi,\pi]$ and we set $\lambda=0.25$. The cumulative cost obeys
\begin{equation}
D(i,j)=c(i,j)+\min\{D(i-1,j),\,D(i,j-1),\,D(i-1,j-1)\},
\end{equation}
with monotonicity and continuity constraints; the complexity is $\mathcal{O}(NW)$. To improve stability, we seed anchors by forcing matches on stationary segments where both speeds are below $0.1\,\mathrm{m/s}$ for at least $2$\,s. The optimal warping path $\mathcal{W}=\{(i_k,j_k)\}_{k=1}^K$ yields a piecewise-linear mapping between timestamps $(t_L[i_k], t_I[j_k])$, from which unified timestamps are obtained via linear interpolation. The resulting time-aligned sequences are then used for GNSS/IMU anchoring and LiDAR registration downstream.
}




\vspace{-1mm}
\subsection{High-Precision 3D LiDAR Map Construction}

\begin{figure}[tb!]
    \centering
    \vspace{-2mm}
    \includegraphics[width=\columnwidth]{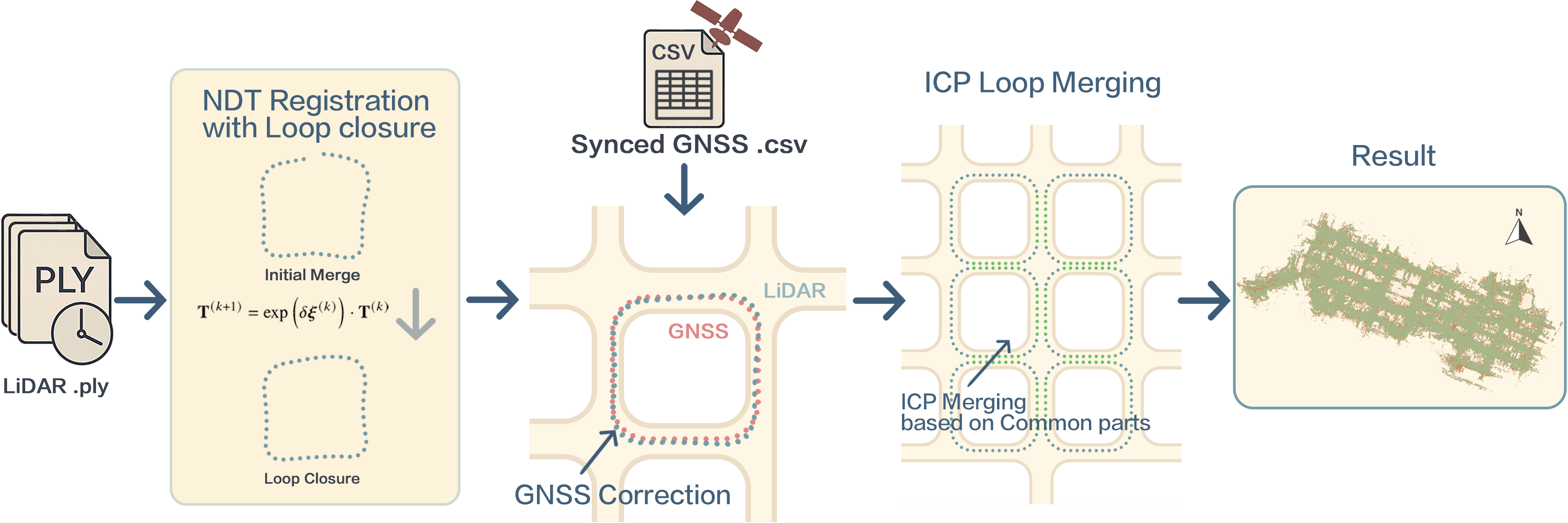}
    \caption{Overview of the proposed mapping pipeline. The process begins with sequential LiDAR frames, followed by NDT-based loop registration, GNSS-informed global calibration, and ICP merging to generate a consistent 3D urban point cloud map.}
    \label{fig:map-construction-pipeline}
    \vspace{-6mm}
\end{figure}

Figure~\ref{fig:map-construction-pipeline} illustrates the proposed mapping pipeline. It comprises three main stages: (1) loop registration using the NDT algorithm and pose graph optimization with loop closures; (2) GNSS-based calibration to anchor local trajectories to geographic coordinates; and (3) global merging of loop segments via ICP alignment on overlapping regions. {Since ICP can be unreliable under weak initialization, small overlap, dynamic objects and repeated structures, we invoke ICP only as a refinement step after GNSS-based anchoring.  Moreover, we apply strict overlap validation to ensure that there is sufficient overlap between the two point clouds. We also perform robust correspondence trimming using distance gating and geometric consistency checks to reduce sensitivity to moving objects and outliers. Finally, in case of a failed refinement, our method performs re-initialization using  global registration on feature correspondences and then retries the refinement, providing a practical recovery mechanism in difficult overlap conditions.} Each stage is detailed in the following subsections.

\subsubsection{Loop Registration and Pose Graph Optimization}
Within each survey loop, sequential LiDAR scans are aligned using NDT~\cite{zaganidis2017semantic}, which models voxelized regions of the reference scan as Gaussian distributions with mean $\boldsymbol{\mu}_j$ and covariance $\boldsymbol{\Sigma}_j$. The rigid-body transformation $\mathbf{T}$ maps each point $\mathbf{p}_i$ from the current scan to its corresponding voxel by maximizing the alignment likelihood $J(\mathbf{T})$.

\vspace{-2mm}
\begin{equation}
J(\mathbf{T}) = -\sum_{i=1}^{N}\exp\left(-\frac{1}{2}(\mathbf{T}\mathbf{p}_i - \mathbf{\mu}_j)^\top \Sigma_j^{-1}(\mathbf{T}\mathbf{p}_i - \mathbf{\mu}_j)\right)
\end{equation}

To maintain global consistency, a pose graph is constructed where edges $\mathcal{E}$ represent relative transformations from NDT-based sequential alignment and loop closures identified via Scan Context descriptors. The optimization minimizes pose discrepancies by solving $\hat{\mathbf{X}} = \arg\min_{\mathbf{X}} \sum_{(i,j)\in \mathcal{E}} \left\| \log\left( \mathbf{Z}_{ij}^{-1} \mathbf{X}_i^{-1} \mathbf{X}_j \right) \right\|_{\Omega_{ij}}^2$, where $\mathbf{X}$ is the set of estimated node poses, $\mathbf{Z}_{ij}$ the observed relative transformation between nodes $i$ and $j$, and $\Omega_{ij}$ the information matrix encoding measurement uncertainty.

\subsubsection{GNSS-Based Loop Calibration}
Following pose graph optimization, residual global drift may persist across loops. To mitigate this, each optimized trajectory $\mathbf{T}_{\text{loop}}$ is aligned to its corresponding GNSS path $\mathbf{G}_{\text{loop}}$ using a weighted least squares approach:
\vspace{-3mm}
\begin{equation}
\mathbf{T}_\text{loop}^{\ast} = \arg\min_{\mathbf{T}}\sum_{i=1}^{M}\left\|\mathbf{G}_\text{loop}^{(i)} - \mathbf{T}\mathbf{T}_\text{loop}^{(i)}\right\|_{\mathbf{W}_{i}}^2
\vspace{2mm}
\end{equation}
Here, $\mathbf{W}_i$ is a diagonal weighting matrix that reflects the quality of the GNSS signal, including factors such as position dilution of precision (PDOP) and the status of the RTK fix. This step anchors local maps to real-world coordinates while de-emphasizing GNSS-denied segments with low weight, thereby preserving local loop integrity. {When a $3\times3$ GNSS position covariance $\boldsymbol{\Sigma}_i$ is available, we set $\mathbf{W}_i=\bigl(\boldsymbol{\Sigma}_i+\sigma_0^2\mathbf{I}_3\bigr)^{-1}$ with $\sigma_0=0.05\,\text{m}$ as a stability floor. If covariance is unavailable, we use a mode-dependent diagonal model $\mathbf{W}_i=\operatorname{diag}(\sigma_{E,i}^{-2},\sigma_{N,i}^{-2},\sigma_{U,i}^{-2})$ with $(\sigma_{E,i},\sigma_{N,i},\sigma_{U,i})=(0.03,0.03,0.05)\,\text{m}$ for RTK~FIX and $(0.50,0.50,1.00)\,\text{m}$ otherwise. Epochs with PDOP$>6$ or missing RTK status are discarded.}

{GNSS samples $\mathbf g_i$ are time-aligned to $\mathbf p_i$ by linearly interpolating the GNSS stream and pairing each pose with the nearest timestamp within $50$\,ms; pairs outside this tolerance are discarded. In this anchoring step we align only the translation $\mathbf p_i$ and keep the loop’s internal rotations unchanged to preserve local loop shape.}

{For the least squares representation, let $\mathbf{p}_i\in\mathbb{R}^3$ be the translation of $\mathbf{T}_\text{loop}^{(i)}$ and $\mathbf{g}_i\in\mathbb{R}^3$ the GNSS position. The problem is}
{
\begin{equation}
(\hat{\mathbf{R}},\hat{\mathbf{t}})=\arg\min_{\mathbf{R}\in SO(3),\,\mathbf{t}\in\mathbb{R}^3}\sum_{i=1}^{M}\bigl\|\mathbf{g}_i-\mathbf{R}\mathbf{p}_i-\mathbf{t}\bigr\|_{\mathbf{W}_i}^{2}.
\end{equation}
{With whitening matrices $\mathbf{L}_i$ such that $\mathbf{L}_i^\top\mathbf{L}_i=\mathbf{W}_i$, define $\tilde{\mathbf{g}}_i=\mathbf{L}_i\mathbf{g}_i$ and $\tilde{\mathbf{p}}_i=\mathbf{L}_i\mathbf{p}_i$, subtract centroids $\bar{\mathbf{g}}$ and $\bar{\mathbf{p}}$, and form $\mathbf{S}=\sum_i(\tilde{\mathbf{p}}_i-\bar{\mathbf{p}})(\tilde{\mathbf{g}}_i-\bar{\mathbf{g}})^\top$. Centroids and the cross-covariance are computed in the whitened space, i.e., $\bar{\tilde{\mathbf g}}=\frac{1}{M}\sum_i \tilde{\mathbf g}_i$, $\bar{\tilde{\mathbf p}}=\frac{1}{M}\sum_i \tilde{\mathbf p}_i$, and $\mathbf S=\sum_i(\tilde{\mathbf p}_i-\bar{\tilde{\mathbf p}})(\tilde{\mathbf g}_i-\bar{\tilde{\mathbf g}})^\top$; when $\mathbf W_i=w_i\mathbf I_3$ this reduces to using weighted centroids in the original space.
If $\mathbf{U}\boldsymbol{\Sigma}\mathbf{V}^\top=\operatorname{svd}(\mathbf{S})$, then $\hat{\mathbf{R}}=\mathbf{V}\operatorname{diag}(1,1,\det(\mathbf{V}\mathbf{U}^\top))\mathbf{U}^\top$ and $\hat{\mathbf{t}}=\bar{\mathbf{g}}-\hat{\mathbf{R}}\bar{\mathbf{p}}$.}
}
{To suppress occasional GNSS outliers we wrap the update in an iteratively reweighted least squares step with a Huber loss ($\delta=1.0$\,m) on each whitened residual.}

\subsubsection{Global Merging Using ICP on Overlapping Regions}
Final map integration is performed by aligning adjacent loops using ICP on overlapping point cloud regions. Fast Point Feature Histograms (FPFH) guide initial correspondence estimation, while Octree-based indexing identifies candidate overlaps. {To address ICP sensitivity to initialization and local minima, we do not start ICP from an identity guess. Instead, the initial inter-loop transform $\mathbf{T}_0$ is obtained from the GNSS-anchored loop frames after Sec.~(GNSS-Based Loop Calibration), which provides a bounded prior for the relative pose. We further restrict ICP to verified overlap windows and use a robust point-to-plane trimmed objective. Specifically, we apply (i) a maximum correspondence distance of $1.0$\,m, (ii) a normal-consistency gate with $\angle(\mathbf{n}_a,\mathbf{n}_b)<30^\circ$, and (iii) trimming ratio $\rho=0.7$ that discards the highest $30\%$ point-pair residuals. A merge is accepted only if the inlier ratio exceeds $0.30$ and the mean inlier residual is below $0.15$\,m. If these criteria are not met, we re-initialize using a robust global registration solver (Go-ICP~\cite{yang2015go}) on the FPFH correspondences and then rerun the refinement. Merges that still fail are rejected to prevent contaminating the global map.}

The alignment process minimizes the mean squared error between corresponding point pairs $(\mathbf{a}_k, \mathbf{b}_k)$ from each loop. This is formalized by the cost function
$E(\mathbf{R}, \mathbf{t}) = \frac{1}{K} \sum_{k=1}^{K} \left\| \mathbf{R} \mathbf{a}_k + \mathbf{t} - \mathbf{b}_k \right\|^2$,
where $\mathbf{R}$ and $\mathbf{t}$ denote the rotation and translation applied to the source point cloud. To improve robustness in dynamic environments (e.g., moving vehicles, pedestrians), a Trimmed ICP variant discards high-residual point pairs. Post-alignment, the map is refined using voxel grid downsampling (0.05~m resolution) and statistical outlier removal, enhancing compactness and structural clarity. The final map provides a reliable basis for downstream use.

\begin{figure}[!t]
    \centering
    \vspace{-4mm}
    \includegraphics[width=\columnwidth]{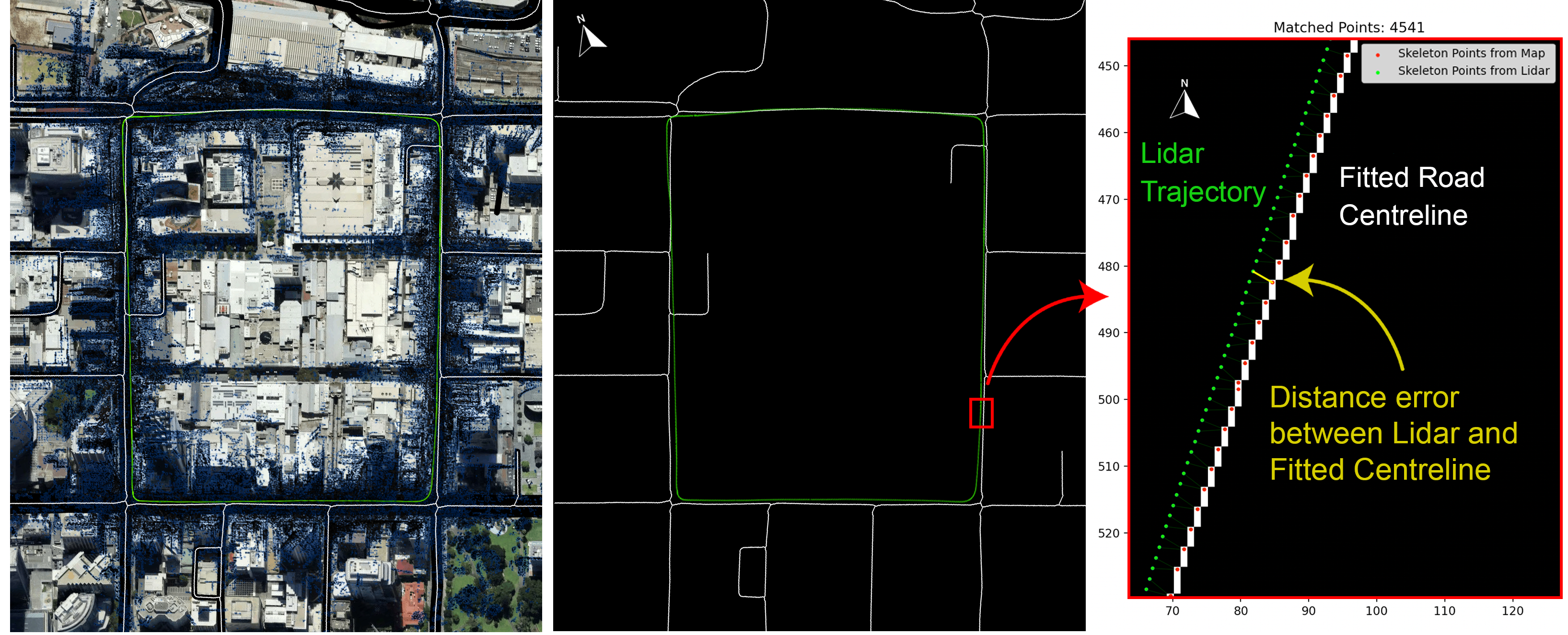}
    \vspace{-6mm}
    \caption{Centreline alignment error analysis. Left: Extracted skeleton overlaid on the LiDAR map. Middle: Zoomed view showing point correspondences between the reference (white) and LiDAR-derived (green) skeletons. Right: Error vectors indicating local geometric deviations due to drift or intersection distortion.}
    \label{fig:error-calculation}
     \vspace{-5mm}
\end{figure}

\vspace{-2mm}
\subsection{Three-Stage Assessment of 3D Map Alignment Accuracy}
To assess the spatial accuracy of the proposed LIGMA framework, we implement a structured three-stage protocol. 
First, road centrelines are extracted from the reconstructed 3D LiDAR map and a georeferenced 2D base map from Google Maps, then compared to evaluate geometric alignment. Next, geometric alignment is assessed by computing point-wise correspondences between the LiDAR-derived and reference centrelines, with intersection regions excluded due to ambiguity in lane geometry and resolution. Lastly, intersection zones are analyzed separately using localized deviation metrics to capture alignment accuracy in these geometrically complex areas. These evaluation protocols are described in the following subsections.

\subsubsection{Extraction of Road Centrelines}
To enable consistent comparison, road centrelines are extracted from both the reconstructed 3D map and a georeferenced 2D base map using a unified skeletonization pipeline. For LiDAR-derived centrelines, ground points are first filtered via voxel downsampling and statistical outlier removal, then projected onto the XY-plane to generate a dense 2D raster. The intensity at each pixel location $(u,v)$ is computed as a Gaussian-weighted sum over all projected points $\mathcal{P}$, given by $I(u,v) = \sum_{p \in \mathcal{P}} \exp\left(-\frac{(x_p - u)^2 + (y_p - v)^2}{2\sigma^2}\right)$, where $\sigma$ controls the spatial influence radius. An adaptive local threshold $T_{\text{local}}(u,v)$ is then applied to produce a binary road mask, defined as $B(u,v) = \mathbf{1}(I(u,v) > T_{\text{local}}(u,v))$, where $\mathbf{1}(\cdot)$ is the indicator function. The resulting binary mask is refined through morphological thinning to yield a one-pixel-wide skeleton $S_{\text{LiDAR}}$. To reduce noise and spurious branches, a pruning process removes short, low-curvature segments as:\vspace{-2mm}
\begin{equation}
S_{\text{LiDAR}}^{\ast} = S_{\text{LiDAR}} \setminus \left\{ b \mid L(b) < L_{\text{min}},\; \kappa(b) < \kappa_{\text{th}} \right\},
\end{equation}
where $L(b)$ is the branch length, $\kappa(b)$ is its local curvature, and $L_{\text{min}}$, $\kappa_{\text{th}}$ are empirically defined thresholds.

\begin{figure}[!t]
\centering
\vspace{-4mm}
\includegraphics[width=\columnwidth]{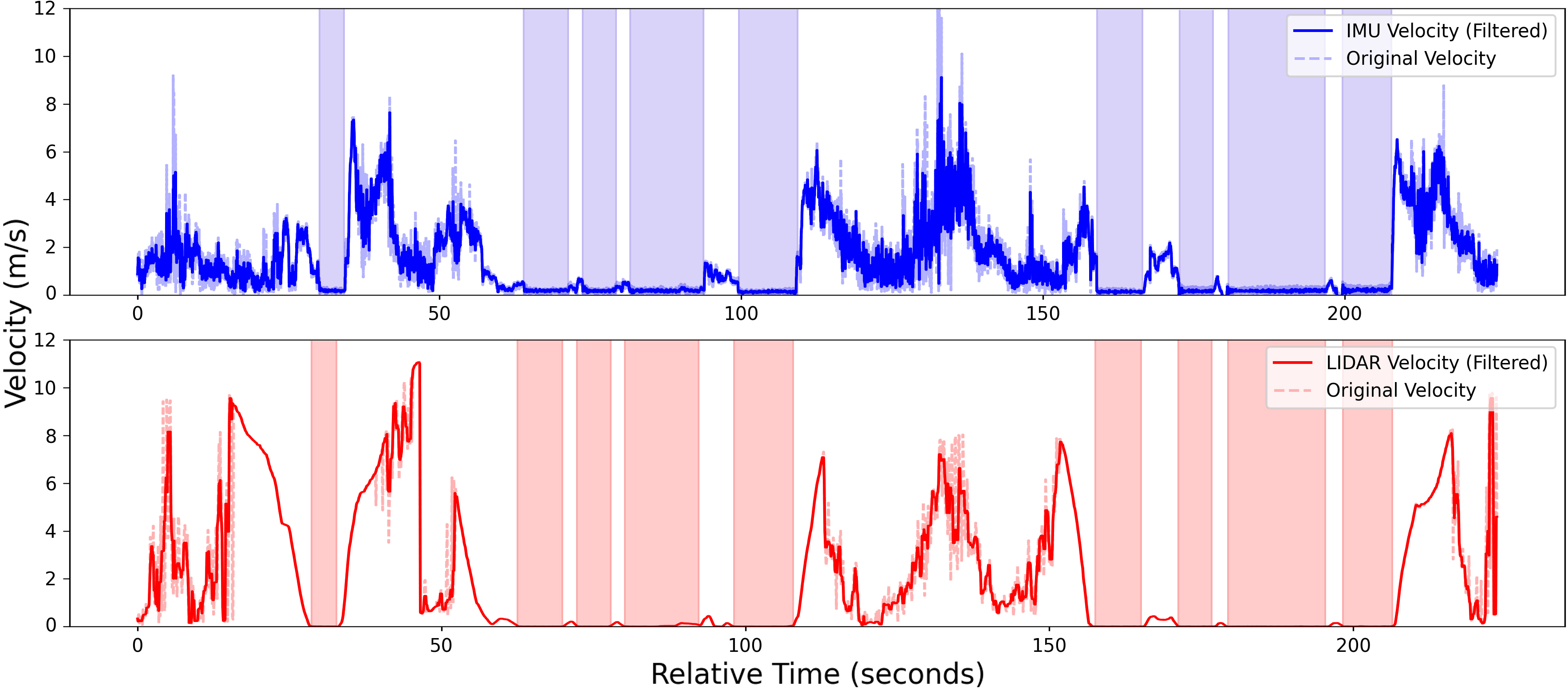}
\vspace{-6mm}
\caption{Velocity profiles from IMU (top, blue) and LiDAR (bottom, red) before DTW. Shaded regions indicate detected stationary phases. Misalignment across modalities underscores the need for temporal correction.}
\vspace{-2mm}
\label{fig:sync-before}
\end{figure}

\begin{figure}[!t]
\centering
\vspace{-2mm}
\includegraphics[width=\columnwidth]{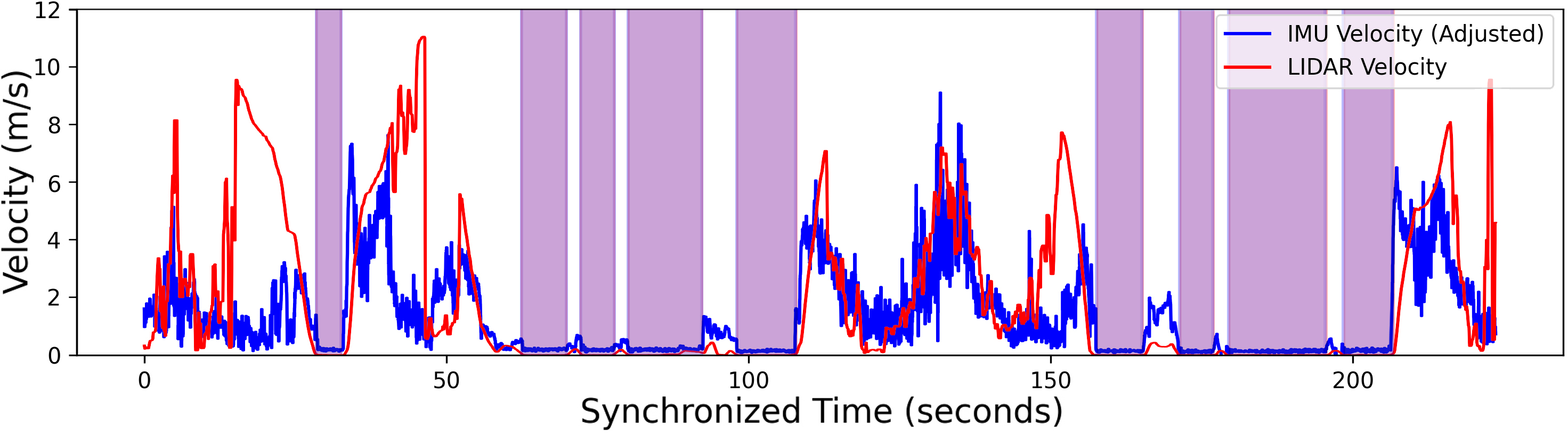}
\vspace{-6mm}
\caption{Synchronized velocity profiles using DTW alignment. The computed time offset ($1.45$ seconds) yields tightly aligned motion and stationary phases across sensors.}
\vspace{-4mm}
\label{fig:sync-after}
\end{figure}

\subsubsection{Centreline-Based Alignment Error Metrics}
To quantify global alignment accuracy, point-wise correspondences are computed between the LiDAR-derived skeleton $S_{\text{LiDAR}}^{\ast}$ and the reference skeleton $S_{\text{2D}}^{\ast}$ using a KD-tree nearest-neighbor search. To avoid distortions from ambiguous road geometries near intersections, all skeleton points located within a 20-meter radius of identified intersections are excluded. This ensures that the alignment evaluation reflects true geometric consistency along unambiguous road segments, unaffected by local drift, occlusion, or topology variation.

For each valid point $p_i \in S_{\text{LiDAR}}^{\ast}$, its alignment error is calculated as the minimum Euclidean distance from its closest counterpart $q_j \in S_{\text{2D}}^{\ast}$, calculated by $d_i = \min_{q_j \in S_{\text{2D}}^{\ast}} \sqrt{(x_{p_i} - x_{q_j})^2 + (y_{p_i} - y_{q_j})^2}$. The distribution of these errors across all matched points is then aggregated using RMSE, offering a statistically robust indicator of global spatial alignment quality. This approach enables an accurate assessment of 3D map alignment quality, with visual evidence illustrated in Figure~\ref{fig:error-calculation}.

\begin{figure*}[t!]
    \centering
    \includegraphics[width=\textwidth]{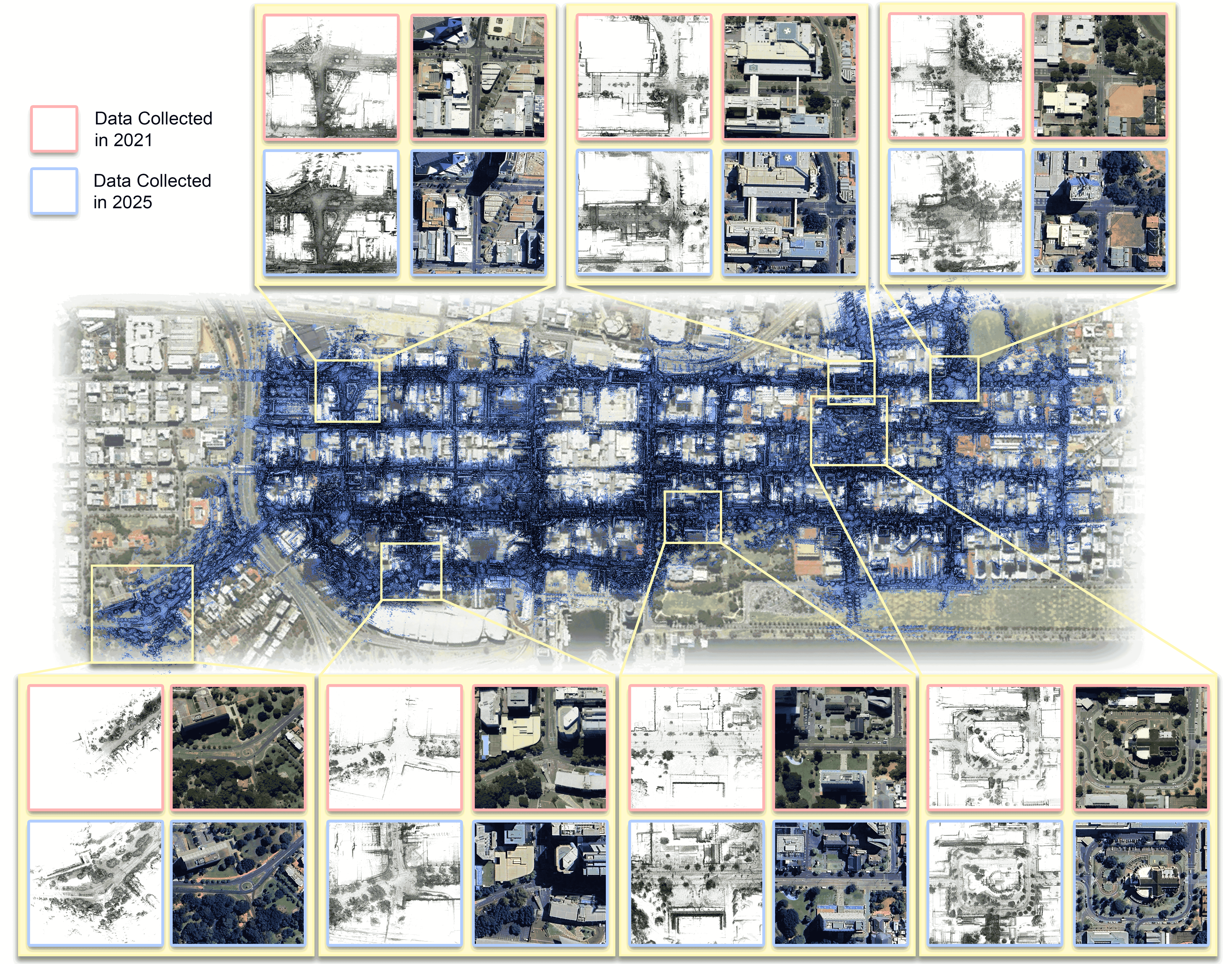}
    \vspace{-2mm}
    \caption{3D LiDAR maps of Perth CBD generated using 2021 OS1-64 (top, red) and 2025 OS1-128 (bottom, blue) data. Insets highlight urban regions overlaid with satellite imagery. The 2025 reconstruction exhibits improved point density, geometric fidelity, and map completeness, demonstrating the advantages of higher-resolution sensing and the proposed framework.}
    \vspace{-4mm}
    \label{fig:map-construction-result}
\end{figure*}

\begin{figure*}[tb!]
\centering
\vspace{-4mm}
\includegraphics[width=\textwidth]{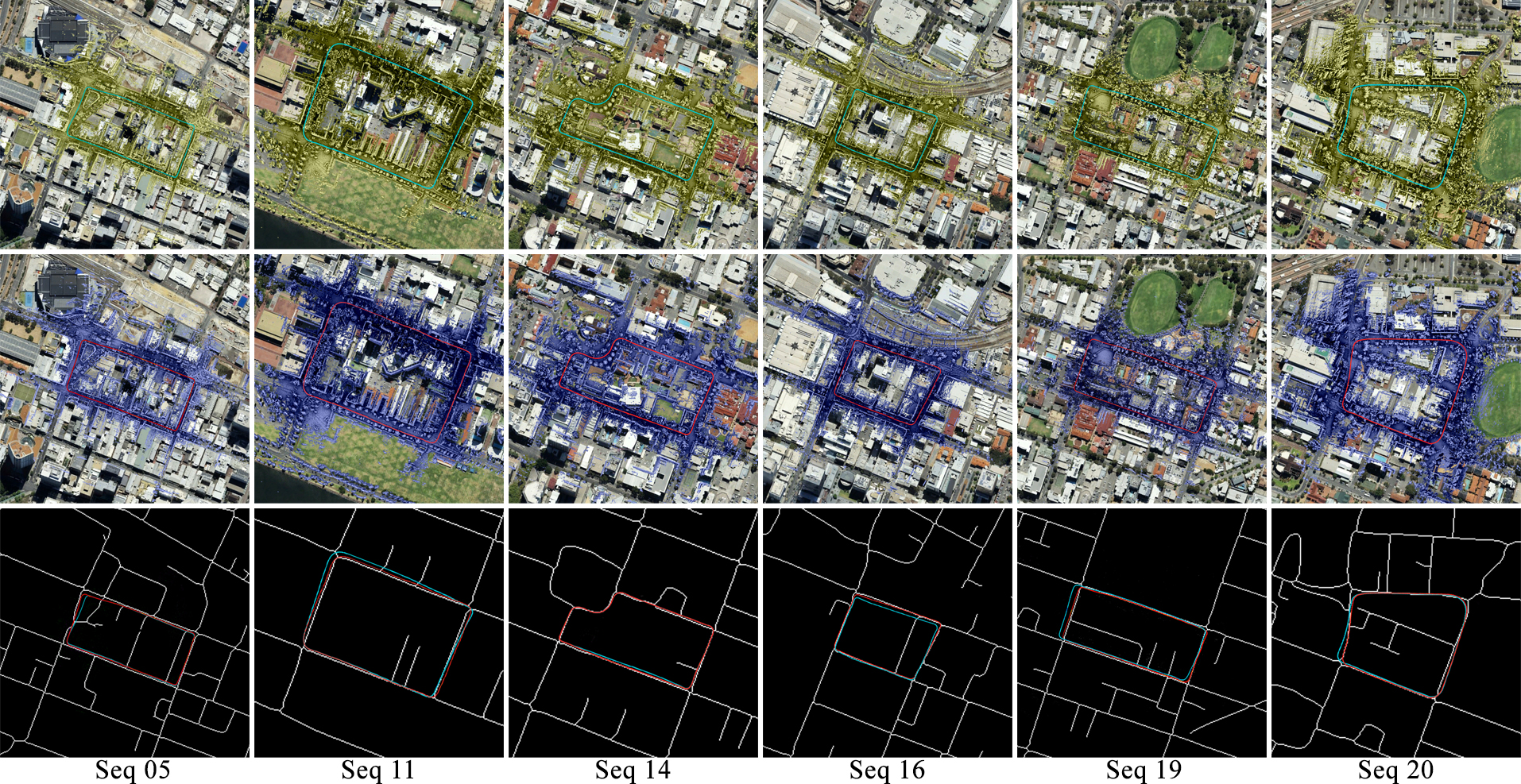}
\vspace{-6mm}
\caption{Qualitative comparison across six representative loops in Perth CBD (Seq~05, Seq~11, Seq~14, Seq~16, Seq~19, Seq~20). \emph{Top row}: LiDAR-only reconstructions over satellite imagery (green). \emph{Middle row}: reconstructions produced by the proposed LIGMA framework over the same areas (blue). \emph{Bottom row}: centreline overlays showing the georeferenced reference (white), the LiDAR-only result (green), and LIGMA (red). The proposed method improves loop closure and intersection geometry while reducing drift and enhancing global consistency.}
\label{fig:loop-uolgim}
\vspace{-7mm}
\end{figure*}

\begin{figure}[tb!]
    \centering
    \includegraphics[width=\columnwidth]{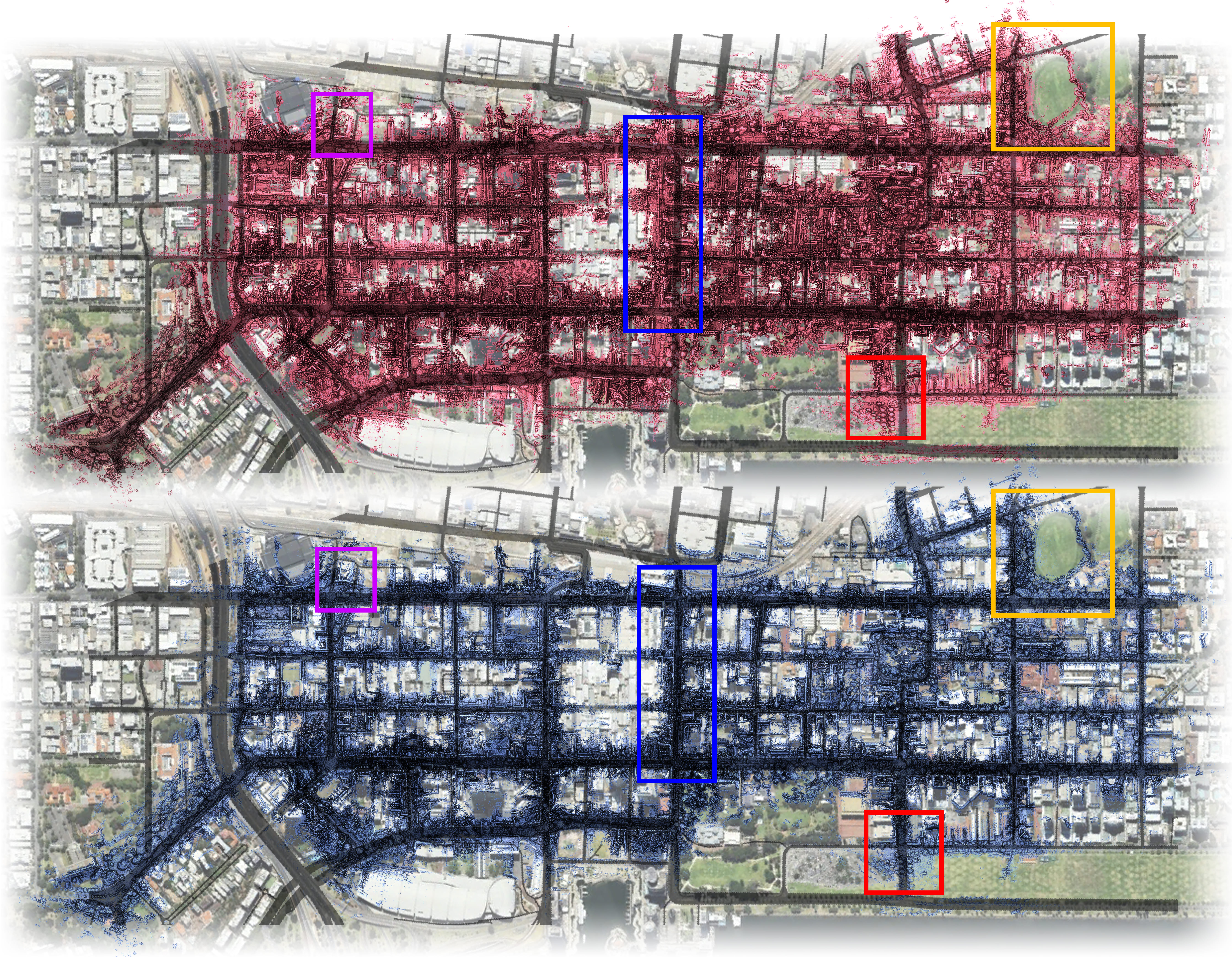}
    \vspace{-6mm}
    \caption{Comparison of 3D urban maps constructed without GNSS calibration (top, red) and with LIGMA (bottom, blue), overlaid on satellite imagery of Perth CBD. Highlighted regions show improved alignment and structural completeness following GNSS/IMU-constrained optimization.}
    \label{fig:u-olgim-comparison}
    \vspace{-8mm}
\end{figure}

\subsubsection{Intersection-Based Local Geometric Consistency Metrics}
To assess the geometric fidelity of reconstructed 3D maps at critical junctions, we introduce a dual-metric approach that evaluates both spatial displacement and directional alignment between corresponding intersections. This method captures not only positional accuracy but also local topological orientation. Let $\mathcal{P}_I = \{p_i, \theta_i\}_{i=1}^{K}$ and $\mathcal{M}_I = \{m_i, \phi_i\}_{i=1}^{K}$ represent intersection centroids and their dominant road orientations from the LiDAR-derived and reference skeletons, respectively. Correspondences are established using mutual nearest-neighbor search within a spatial threshold $\delta$, while preserving node degree and connectivity. The \textit{spatial offset metric} $\mathcal{E}_{\text{spatial}}$ jointly penalizes positional and angular mismatches:
\vspace{-3mm}
\begin{equation}
\mathcal{E}_{\text{spatial}} = \frac{1}{K} \sum_{i=1}^{K} \left[ \alpha \|p_i - m_i\|_2 + \beta \cdot d_{\mathcal{T}}(p_i, m_i) \right]
\end{equation}
Here, $\alpha$ and $\beta$ are weights, and $d_{\mathcal{T}}(p_i, m_i)$ quantifies the angular deviation between the outgoing road branches. For each intersection \( i \), it is defined as \( d_{\mathcal{T}}(p_i, m_i) = \frac{1}{D_i} \sum_{j=1}^{D_i} \min_k \left| \theta_{i,j} - \phi_{i,k} \right| \), where \( D_i \) is the number of outgoing branches at \( p_i \), \( \theta_{i,j} \) is the orientation of the \( j \)-th branch in the LiDAR-derived map, and \( \phi_{i,k} \) is the most closely aligned direction in the reference map. This ensures optimal directional pairing between the corresponding intersections. To quantify local shape distortion, we define the \textit{envelope error} $\mathcal{E}_{\text{env}}$ as a metric that captures the worst-case spatial deviation between neighborhoods around each matched intersection:
\vspace{-3mm}
\begin{equation}
\mathcal{E}_{\text{env}} = \frac{1}{K} \sum_{i=1}^{K} \max_{x \in \mathcal{B}(p_i, r)} \min_{y \in \mathcal{B}(m_i, r)} \| x - y \|_2
\end{equation}

Here, $\mathcal{B}(p_i, r)$ and $\mathcal{B}(m_i, r)$ denote the sets of skeleton points within radius $r$ centered at $p_i$ and $m_i$, respectively. Together with the spatial offset metric, this newly defined envelope error offers a robust, topology-aware evaluation of intersection alignment, capturing localized drift and deformation not reflected in global skeleton RMSE (see Fig.~\ref{fig:error-calculation}).

\vspace{-1mm}
\section{Results}
\label{Results}
\vspace{-2mm}

This section evaluates the performance of the proposed framework through a {five-stage} analysis. First, the experimental setup is outlined to support reproducibility. Second, sensor synchronization accuracy is examined, with emphasis on temporal alignment across LiDAR, GNSS, and IMU streams. Third, the results of 3D map reconstruction for Perth CBD are benchmarked against LiDAR-only baselines. {Fourth, skeleton-based centreline metrics are employed to assess spatial accuracy and geometric consistency. Fifth, we report the end-to-end runtime and a per-stage breakdown in the Timing subsection.} Each stage is described in detail in the subsequent subsections.

\vspace{-3mm}
\subsection{Experimental Setup}
The experiments were conducted on a high-performance workstation equipped with an Intel Xeon Gold 6330 CPU (28 cores), 512\,GB of RAM, and an NVIDIA RTX 4090 GPU, running Ubuntu 20.04 LTS. To ensure reproducibility, all software components were containerized using Docker. The data processing pipeline utilized the Ouster SDK v3.1.1 for LiDAR decoding, Open3D v0.17 for point cloud filtering, and the Inertial Sense uINS SDK v2.15 for GNSS–IMU data handling. Pose graph optimization was performed using GTSAM 4.2, while NDT and ICP registration were implemented via the Point Cloud Library (PCL) v1.13.

{On the 2025 Perth–CBD run (21 loops), end-to-end processing excluding acquisition took 12\,h\,44\,m on our workstation; the synchronization step accounted for 243\,m, loop registration 195\,m, GNSS anchoring 126\,m, and loop merging 131\,m. Because our target is offline high-fidelity city-scale mapping rather than real-time deployment, we neither optimize for latency or memory nor derive formal time- or space-complexity bounds beyond the DTW order and this framing keeps the contribution centered on accurate map construction.}

\begin{figure*}[t!]
    \centering
    \begin{minipage}[t]{0.48\textwidth}
        \centering
        \includegraphics[width=\linewidth]{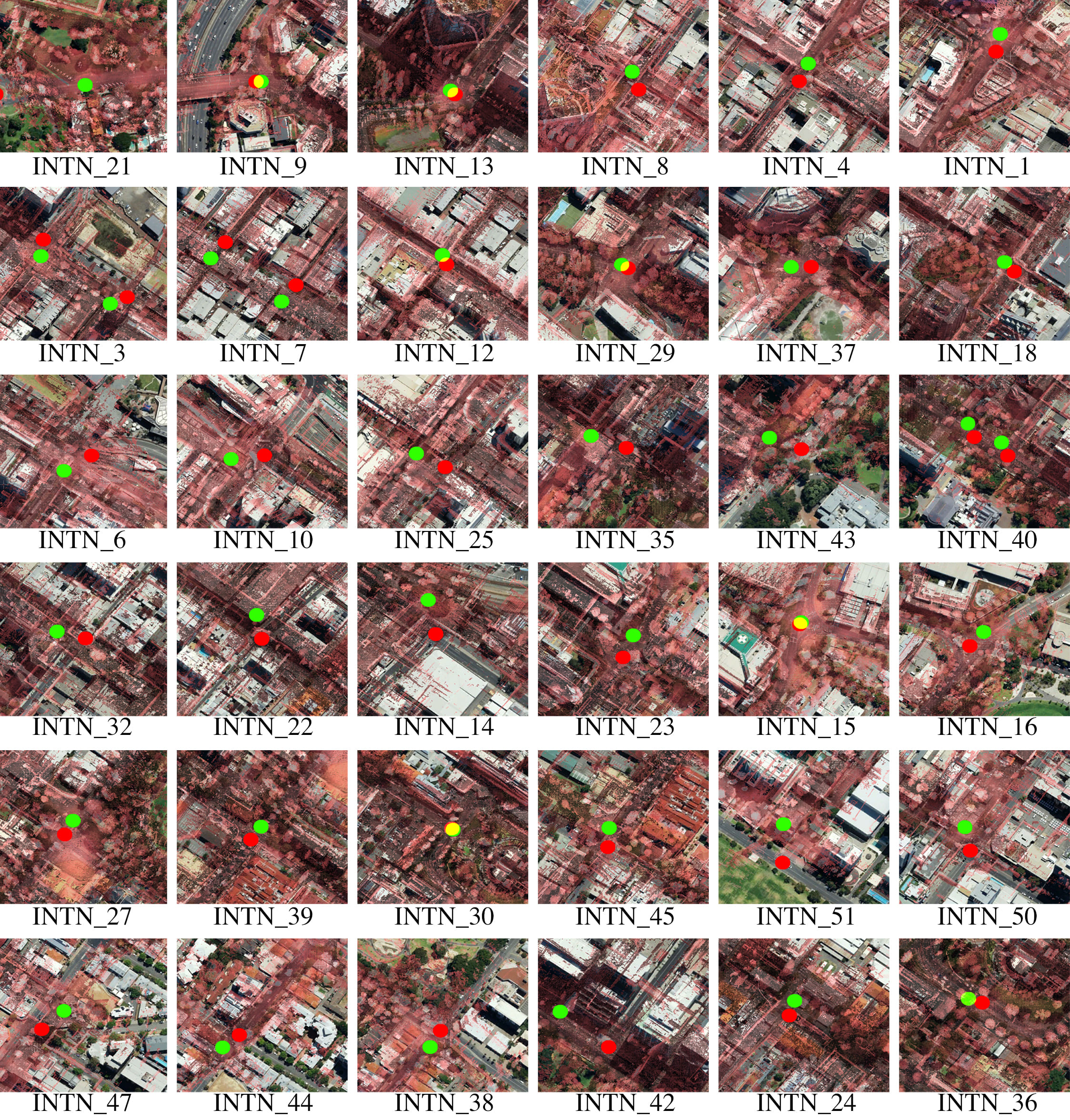}
        \vspace{-5mm}
        \caption{Intersection centroid alignment using the LiDAR-only baseline. 
        Green points represent the reference centroids and red points denote the LiDAR-only results. 
        Yellow regions highlight areas of overlap between the two, emphasizing the degree of alignment.}
        \label{fig:intersection-NDT}
        \vspace{-3mm}
    \end{minipage}
    \hfill
    \begin{minipage}[t]{0.48\textwidth}
        \centering
        \includegraphics[width=\linewidth]{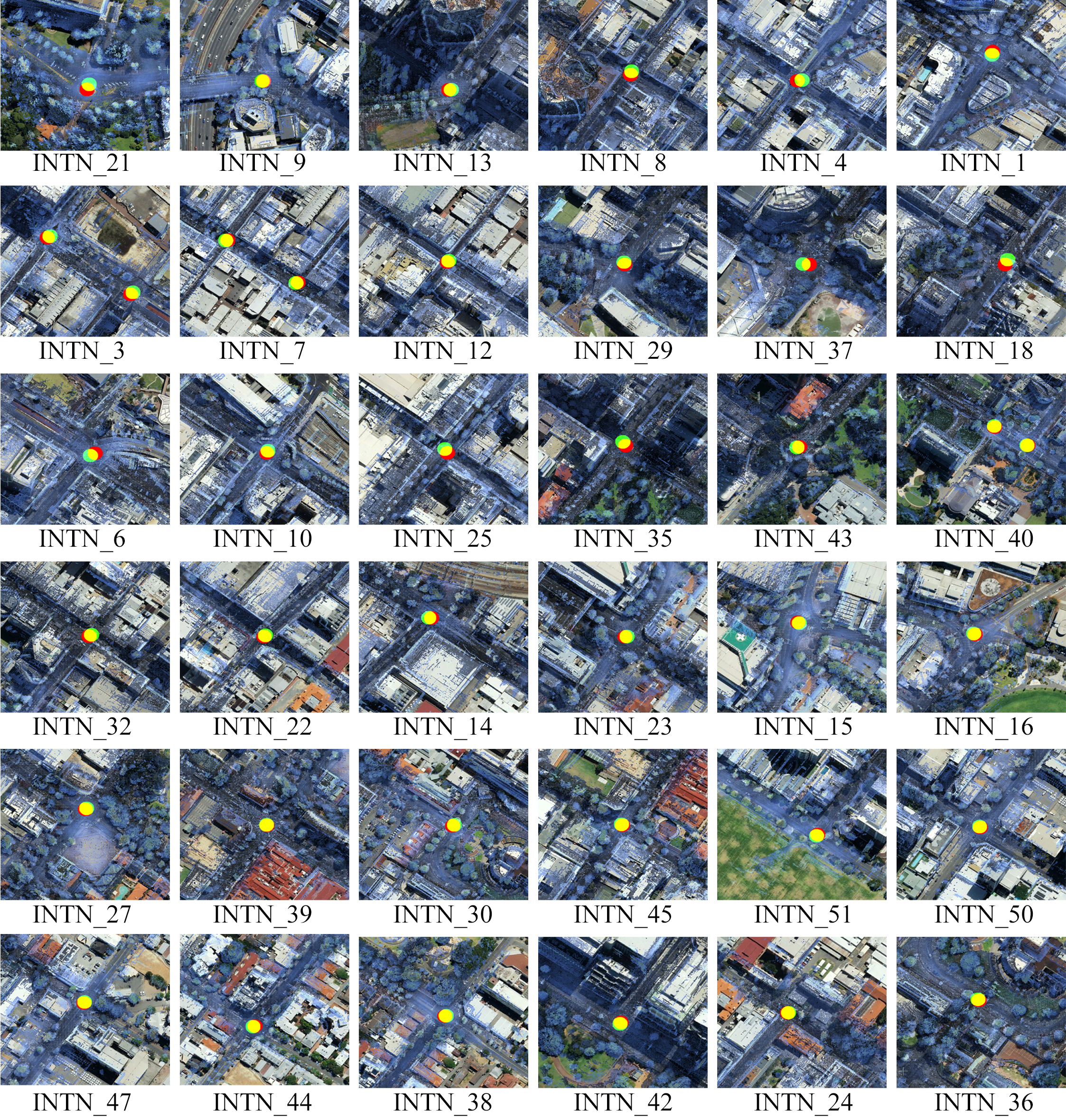}
        \vspace{-5mm}
        \caption{Intersection centroid alignment using the proposed LIGMA framework. 
        Green points represent the reference centroids and red points denote the fused results. 
        Yellow regions indicate overlapping areas between the two, providing a clearer visualization of improved correspondence.}
        \label{fig:intersection-U}
    \end{minipage}
    \vspace{-5mm}
\end{figure*}

\vspace{-3mm}
\subsection{Effectiveness of Multimodal Sensor Data Synchronization}
We evaluated the proposed synchronization framework by comparing IMU and LiDAR velocity profiles before and after alignment. Accurate temporal calibration is critical for sensor fusion, especially for pose estimation and 3D mapping. Despite hardware-level synchronization via PPS triggers, residual time offsets often persist due to jitter, latency, or clock drift, which results in misaligned motion events, most evident in dynamic urban environments. As shown in Figure~\ref{fig:sync-before}, pre-alignment profiles display mismatched transitions and inconsistent stationary phases, degrading point cloud registration and pose accuracy. To address this, we applied a DTW-based velocity matching method, which automatically estimated a $1.45\,\mathrm{s}$ offset. Post-synchronization results (Figure~\ref{fig:sync-after}) demonstrate close alignment of both static and dynamic phases, effectively synchronizing acceleration and deceleration events across sensors. 

{
Quantifying residuals and comparing to a constant-shift baseline. With PPS-only configuration, the median absolute timing error was 18.7\,ms (p95 42.3\,ms). A constant shift estimated via cross-correlation reduced the speed-profile RMSE from 0.41 to 0.22\,m/s and increased the Pearson correlation from 0.72 to 0.85. DTW further reduced RMSE to 0.15\,m/s and increased correlation to 0.93. The p95 residual timing drift, computed from the inferred mapping $t\mapsto t+\delta(t)$, decreased from 42.3\,ms (PPS-only) to 21.4\,ms (constant shift) and to 7.8\,ms (DTW). Table~\ref{tab:sync_ablation} summarizes results over all 21 sequences. At the mapping level, replacing DTW with a constant shift increased the average centreline RMSE from 1.24\,m to 1.38\,m.
}

\begin{table}[t]
\centering
\vspace{-2mm}
\caption{{Synchronization accuracy across 21 sequences. Corr: Pearson correlation between speed profiles. RMSE: speed error. p95 drift: 95th percentile of $|\delta(t)-\mathrm{median}(\delta)|$.}}
\vspace{-3mm}
\label{tab:sync_ablation}
\renewcommand{\arraystretch}{1.1}
          
\begin{tabular}{lccc}
\toprule
Method & Corr $\uparrow$ & RMSE (m/s) $\downarrow$ & p95 drift (ms) $\downarrow$ \\
\midrule
PPS only (no correction) & 0.72 & 0.41 & 42.3 \\
Constant shift (XCorr)   & 0.85 & 0.22 & 21.4 \\
DTW (ours)               & 0.93 & 0.15 & 7.8 \\
\bottomrule
\end{tabular}
\vspace{-7mm}
\end{table}

\vspace{-3mm}
\subsection{High-Precision 3D LiDAR Map Construction Results}
The developed LIGMA framework was applied to generate a high-precision 3D point cloud map of Perth CBD using data collected in 2025, and compared against a baseline map constructed in 2021. Figure~\ref{fig:map-construction-result} illustrates representative urban areas with overlaid point clouds from both maps on high-resolution satellite imagery. The baseline map was generated using an OS1-64 LiDAR, whereas the recent map incorporates data from an OS1-128 sensor with full GNSS/IMU fusion. The recent map exhibits denser point coverage, sharper building edges, and improved reconstruction of roads and intersections. In contrast, the baseline map shows lower resolution, structural gaps, and misalignments, particularly in occluded or dynamic areas. These artifacts are substantially reduced in the recent map as a result of GNSS/IMU-constrained optimization, which mitigates drift and enhances global consistency. Figure~\ref{fig:u-olgim-comparison} further highlights these improvements by comparing outputs without GNSS calibration (top, red) and with LIGMA-based calibration (bottom, blue). The calibrated result demonstrates improved alignment and structural coherence across key city-level features. Beyond geometric accuracy, the temporal alignment of the two maps enables the detection of long-term citywide changes, including new buildings, reconfigured public spaces, and updated road layouts. These changes support application infrastructure planning, HD map updates, and metro change detection.

\vspace{-3mm}
\subsection{Comparative Evaluation of Urban 3D Map Alignment}
\label{sec:comparative-evaluation}
To quantitatively assess the improvements introduced by the proposed framework, we conduct a comparative analysis against a baseline LiDAR-only approach (NDT-ICP, without GNSS). This evaluation includes visual comparisons, skeleton-based analysis, and intersection-level assessments.



\subsubsection{Visual Comparison of Mapping Results}
A qualitative comparison is presented between the map generated by the LiDAR-only baseline method and the map produced by the proposed GNSS-calibrated framework. As shown in Fig.~\ref{fig:u-olgim-comparison}, the top row displays the baseline result (red), while the bottom row shows the LIGMA output (blue), both overlaid on satellite imagery. The LiDAR-only approach exhibits noticeable geometric distortions, particularly at intersections and along extended road segments, due to cumulative drift from locally constrained point cloud alignment. In contrast, the proposed method incorporates GNSS calibration and global optimization, yielding improved geometric consistency and reduced drift. These improvements are especially pronounced in dense city layouts, where intersection alignment and structural continuity are markedly enhanced.




\begin{table}
\vspace{-2mm}
\centering
\caption{Comparison of road centreline alignment metrics across all test sequences for the LiDAR-only baseline and the proposed framework. Relative improvements indicate percentage reductions in alignment error.}
\vspace{-2mm}
\label{tab:centreline_metrics_comparison}
\renewcommand{\arraystretch}{1.15}
\resizebox{\columnwidth}{!}{
\begin{tabular}{lcccccccccccc}
\toprule
\multirow{2}{*}{Sequence} & 
\multicolumn{4}{c}{LiDAR Only (No GNSS)} & 
\multicolumn{4}{c}{LIGMA (GNSS-Calibrated)} & 
\multicolumn{4}{c}{Relative Improvement (\%)} \\
\cmidrule(lr){2-5} \cmidrule(lr){6-9} \cmidrule(lr){10-13}
 & RMSE & Min & Max & SD & RMSE & Min & Max & SD & RMSE & Min & Max & SD \\
\midrule
Seq 01 & 2.82 & 0.24 & 6.78 & 1.41 & 1.14 & 0.12 & 2.63 & 0.58 & 59.6 & 50.0 & 61.2 & 58.9 \\
Seq 02 & 3.45 & 0.31 & 7.42 & 1.78 & 1.36 & 0.15 & 3.21 & 0.72 & 60.6 & 51.6 & 56.7 & 59.6 \\
Seq 03 & 2.31 & 0.25 & 5.35 & 1.20 & 0.92 & 0.12 & 2.37 & 0.48 & 60.2 & 52.0 & 55.7 & 60.0 \\
Seq 04 & 3.73 & 0.35 & 8.75 & 1.91 & 1.52 & 0.16 & 3.42 & 0.81 & 59.2 & 54.3 & 60.9 & 57.6 \\
Seq 05 & 2.74 & 0.28 & 6.38 & 1.42 & 1.05 & 0.13 & 2.42 & 0.55 & 61.7 & 53.6 & 62.1 & 61.3 \\
Seq 06 & 4.18 & 0.39 & 9.53 & 2.15 & 1.62 & 0.18 & 3.45 & 0.85 & 61.2 & 53.8 & 63.8 & 60.5 \\
Seq 07 & 2.29 & 0.23 & 5.86 & 1.22 & 0.88 & 0.11 & 2.32 & 0.45 & 61.6 & 52.2 & 60.4 & 63.1 \\
Seq 08 & 3.92 & 0.32 & 8.14 & 1.95 & 1.47 & 0.15 & 3.33 & 0.76 & 62.5 & 53.1 & 59.1 & 61.0 \\
Seq 09 & 2.87 & 0.26 & 6.29 & 1.45 & 1.09 & 0.13 & 2.45 & 0.52 & 62.0 & 50.0 & 61.0 & 64.1 \\
Seq 10 & 3.65 & 0.34 & 8.32 & 1.87 & 1.41 & 0.17 & 3.27 & 0.73 & 61.4 & 50.0 & 60.7 & 61.0 \\
Seq 11 & 3.13 & 0.29 & 6.92 & 1.58 & 1.18 & 0.14 & 2.84 & 0.61 & 62.3 & 51.7 & 59.0 & 61.4 \\
Seq 12 & 4.31 & 0.41 & 9.76 & 2.18 & 1.62 & 0.19 & 3.38 & 0.85 & 62.4 & 53.7 & 65.4 & 61.0 \\
Seq 13 & 2.48 & 0.25 & 5.57 & 1.32 & 0.95 & 0.12 & 2.28 & 0.48 & 61.7 & 52.0 & 59.1 & 63.6 \\
Seq 14 & 3.76 & 0.36 & 8.82 & 1.92 & 1.41 & 0.16 & 3.32 & 0.75 & 62.5 & 55.6 & 62.4 & 60.9 \\
Seq 15 & 2.95 & 0.27 & 6.68 & 1.47 & 1.13 & 0.13 & 2.62 & 0.57 & 61.7 & 51.9 & 60.8 & 61.2 \\
Seq 16 & 3.42 & 0.33 & 8.04 & 1.76 & 1.35 & 0.16 & 3.25 & 0.68 & 60.5 & 51.5 & 59.6 & 61.4 \\
Seq 17 & 3.35 & 0.31 & 7.73 & 1.68 & 1.28 & 0.15 & 3.14 & 0.65 & 61.8 & 51.6 & 59.4 & 61.3 \\
Seq 18 & 4.47 & 0.42 & 9.98 & 2.24 & 1.68 & 0.19 & 3.46 & 0.86 & 62.4 & 54.8 & 65.3 & 61.6 \\
Seq 19 & 2.63 & 0.25 & 6.21 & 1.35 & 1.02 & 0.13 & 2.52 & 0.52 & 61.2 & 48.0 & 59.4 & 61.5 \\
Seq 20 & 3.87 & 0.37 & 8.35 & 1.98 & 1.48 & 0.17 & 3.36 & 0.77 & 61.8 & 54.1 & 59.8 & 61.1 \\
Seq 21 & 4.65 & 0.43 & 10.20 & 2.30 & 1.72 & 0.20 & 3.55 & 0.88 & 63.0 & 55.5 & 66.0 & 62.0 \\
\midrule
\textbf{Average} & \textbf{3.32} & \textbf{0.31} & \textbf{7.54} & \textbf{1.69} & \textbf{1.24} & \textbf{0.15} & \textbf{2.95} & \textbf{0.65} & \textbf{61.4} & \textbf{52.3} & \textbf{60.6} & \textbf{61.1} \\
\bottomrule
\end{tabular}}
\vspace{-7mm}
\end{table}

\subsubsection{Loop-Based Centreline Comparison}

To assess route-level consistency, we selected six representative loops from the twenty-one sequences in the Perth CBD (Seq 05, 11, 14, 16, 19, and 20). These loops span long straight segments, curved roads, and dense junctions. Figure~\ref{fig:loop-uolgim} consolidates the visual evidence: for each loop, the top row shows the LiDAR-only baseline overlaid on satellite imagery, the middle row shows the reconstruction produced by LIGMA, and the bottom row presents the corresponding road centreline comparison against a 2D cartographic reference. The baseline typically exhibits drift-induced shear and irregular loop closures, with noticeable misalignment at intersections and along extended segments. In contrast, LIGMA yields coherent loop geometry and improved intersection alignment, with centrelines closely following the reference. While Fig.~\ref{fig:loop-uolgim} illustrates typical behavior on six loops, the quantitative evaluation over all twenty-one sequences is summarized in Table~\ref{tab:centreline_metrics_comparison}: our framework reduces centreline RMSE from $3.32$\,m (LiDAR-only) to $1.24$\,m on average, with consistent gains in the minimum, maximum, and standard deviation of the alignment error.

\begin{table}
\centering
\caption{Intersection-level alignment metrics for 15 of 51 intersections (subset shown due to space constraints). CI: Centroidal Offset (m), MOD: Mean Orientation Deviation (\degree). Abs. = absolute reduction; Pct. = percentage reduction relative to baseline.}
\vspace{-2mm}
\label{tab:intersection_metrics}
\renewcommand{\arraystretch}{1.08}
\resizebox{\columnwidth}{!}{
\begin{tabular}{cccccccccc}
\toprule
\multirow{2}{*}{Intersection} & 
\multicolumn{2}{c}{LiDAR-only} & 
\multicolumn{2}{c}{LIGMA} & 
\multicolumn{4}{c}{Relative Improvement} \\
\cmidrule(lr){2-3} \cmidrule(lr){4-5} \cmidrule(lr){6-9}
& CI & MOD & CI & MOD & Abs. CI & Abs. MOD & Pct. CI & Pct. MOD \\
\midrule
1  & 11.50 & 2.67 & 2.36 & 0.85 & 9.14 & 1.82 & 79.5 & 68.2 \\
3  & 11.32 & 3.21 & 1.38 & 1.48 & 9.94 & 1.73 & 87.8 & 46.1 \\
7  & 12.68 & 1.94 & 1.32 & 1.23 & 11.36 & 0.71 & 89.6 & 36.6 \\
10 & 14.53 & 3.05 & 1.47 & 0.64 & 13.06 & 2.41 & 89.9 & 79.0 \\
14 & 24.09 & 2.47 & 3.63 & 1.86 & 20.46 & 0.61 & 85.0 & 24.7 \\
21 & 68.73 & 1.89 & 5.73 & 1.73 & 63.00 & 0.16 & 91.8 & 8.5 \\
23 & 17.47 & 3.26 & 0.95 & 0.43 & 16.52 & 2.83 & 94.6 & 86.8 \\
25 & 19.72 & 1.96 & 1.87 & 1.67 & 17.85 & 0.29 & 90.5 & 17.4 \\
27 & 10.73 & 3.41 & 1.02 & 0.58 & 9.71 & 2.83 & 90.5 & 83.0 \\
31 & 21.38 & 1.59 & 4.12 & 0.67 & 17.26 & 0.92 & 80.7 & 57.9 \\
38 & 14.45 & 2.92 & 0.26 & 0.48 & 14.19 & 2.44 & 98.2 & 83.6 \\
42 & 25.05 & 3.38 & 1.30 & 1.23 & 23.75 & 2.15 & 94.8 & 63.6 \\
45 & 15.96 & 4.62 & 0.96 & 0.96 & 15.00 & 3.66 & 94.0 & 79.2 \\
47 & 16.41 & 3.29 & 1.10 & 1.38 & 15.31 & 1.91 & 93.3 & 58.1 \\
51 & 25.48 & 3.47 & 2.26 & 1.31 & 23.22 & 2.16 & 91.1 & 62.2 \\
\bottomrule
\end{tabular}}
\vspace{-3mm}
\end{table}

\subsubsection{Intersection-Based Alignment Evaluation}
The alignment of the intersections provides key insight into the local geometric consistency of the 3D city maps. We assess spatial accuracy by matching centroids extracted from skeletonized LiDAR maps with those from a georeferenced reference. Figures~\ref{fig:intersection-NDT} and \ref{fig:intersection-U} compare the results of the LiDAR-only method and the proposed framework. Green points indicate ground-truth centroids; red points show those extracted from reconstructed maps. The baseline shows clear misalignments, especially at complex or peripheral junctions due to accumulated drift. Our proposed method yields close centroid correspondence, indicating higher alignment accuracy. Alignment is quantified using four metrics: centroid offset (CO, in meters), mean orientation deviation (MOD, in degrees), local envelope error (LEE, maximum deviation within a 10\,m radius), and standard deviation (SD) of centroid offset.

Table~\ref{tab:intersection_metrics} summarizes results for 15 selected intersections, with percentage improvements shown in the right most columns. Table~\ref{tab:alignment_stats_summary} summarizes average centerline and intersection alignment metrics across all test sequences. The proposed approach reduces centroid offset by 61.5\%, orientation deviation by 55.1\%, and LEE by 59.4\% over the baseline. These improvements are most pronounced at challenging junctions and map boundaries. Accurate intersection alignment supports critical tasks such as trajectory planning, and robust localization. These results demonstrate the effectiveness of the proposed GNSS-constrained fusion framework.

\begin{table}
\centering
\vspace{-2mm}
\caption{Summary statistics of centreline and intersection alignment metrics averaged over all test sequences. CI: Centroidal Offset (m); MOD: Mean Orientation Deviation (\degree); RMSE: Root Mean Square Error of centreline deviation (m); SD: Standard Deviation (m). Relative improvements are computed with respect to the LiDAR-only baseline.}
\vspace{-2mm}
\label{tab:alignment_stats_summary}
\renewcommand{\arraystretch}{1.08}
\resizebox{\columnwidth}{!}{
\begin{tabular}{lcccccc}
\toprule
\multirow{2}{*}{Metric} & 
\multicolumn{2}{c}{LiDAR-only} & 
\multicolumn{2}{c}{LIGMA} & 
\multicolumn{2}{c}{Relative Improvement} \\
\cmidrule(lr){2-3} \cmidrule(lr){4-5} \cmidrule(lr){6-7}
 & Mean & SD & Mean & SD & Abs. & Pct. \\
\midrule
Centreline RMSE (m)        & 3.32  & 1.69  & 1.24 & 0.65 & 2.08 & 62.7 \\
Centreline Min (m)         & 0.31  & 0.06  & 0.15 & 0.02 & 0.16 & 51.6 \\
Centreline Max (m)         & 7.54  & 1.35  & 2.95 & 0.35 & 4.59 & 60.9 \\
Intersection CI (m)        & 13.22 & 9.11  & 2.01 & 1.01 & 11.21 & 84.8 \\
Intersection MOD (\degree) & 2.56  & 0.72  & 1.18 & 0.45 & 1.38 & 53.9 \\
\bottomrule
\end{tabular}}
\vspace{-6mm}
\end{table}

\vspace{-2mm}
\subsection{Timing}
The complete 3D map construction took 14 hours and 39 minutes (879 minutes). The complete 21 loops data was collected in one go and took 115 minutes. Preprocessing took 68 minutes. The DTW synchronization comprising seeding, bounded DTW on speed/heading with anchors took 11.57 min/loop. Timestamp range trimming after synchronization took less than a minute. Loop registration using NDT and pose-graph optimization took 9.3 min/loop . GNSS-based loop calibration using weighted rigid alignment took 6 min/loop and finally ICP on overlapping regions combined with pruning and consolidation took 6.24 min/loop. The total time per loop comes out to be 41.87 minutes. We emphasize that our work targets high-fidelity city-scale 3D map construction rather than real-time deployment. Constructing an accurate 3D city map spanning 4.2 km$^2$ in under 15 hours (including data capture) remains a significant achievement. 

\color{black}
\vspace{-3mm}
\section{Conclusion}
\label{Conclusion}
\vspace{-1mm}

This paper presented LIGMA, a unified LiDAR–GNSS–IMU fusion framework for generating high-resolution and globally consistent 3D urban maps. The method is built on our proposed multimodal dataset, collected in Perth CBD, which includes LiDAR scans, GNSS trajectories, and IMU data. The framework addresses key challenges such as drift and cross-sensor misalignment through a tightly integrated pipeline. It incorporates LiDAR denoising, GNSS/IMU fusion using an Extended Kalman Filter, and temporal synchronization via velocity-based Dynamic Time Warping. Hierarchical registration is achieved through NDT-based local alignment, GNSS-constrained pose graph optimization, and ICP refinement, ensuring both local and global geometric consistency. Quantitative evaluation using road skeleton alignment shows a 61.4\% reduction in global registration error compared to a LiDAR-only baseline. The average centreline RMSE decreases from 3.32~m to 1.24~m, while the mean intersection centroidal offset drops from 13.22~m to 2.01~m. Mean orientation deviation at intersections is also reduced from 2.56\degree{} to 1.18\degree{}. These improvements confirm that our method delivers substantial gains in both global and local alignment accuracy. The framework supports a range of downstream applications, including autonomous navigation, HD map generation, metropolitan planning, and infrastructure monitoring. Future work will focus on extending the system to operate in GNSS-denied or highly dynamic environments, with adaptations for real-time processing and improved robustness under occlusion and sensor degradation.

\vspace{-3mm}
\section*{Acknowledgment}
\label{Acknowledgment}
\vspace{-1mm}

\small{
This work was funded by the University of Jeddah, Jeddah, Saudi Arabia, under grant No.(UJ-24-SUTU-1290). The authors, therefore, thank the University of Jeddah for its technical and financial support. This project is a collaborative effort between the University of Jeddah and the University of Western Australia. The authors also extend their appreciation to Mangoes Mapping Pty Ltd for their in-kind contribution of data collection equipment, including LiDAR and GNSS systems.
}

\bibliographystyle{IEEEtran}
\bibliography{main}

\end{document}